\pdfoutput=1
\documentclass[11pt]{article}

\usepackage{EMNLP2022}

\usepackage{times}
\usepackage{latexsym}

\usepackage[T1]{fontenc}
\usepackage[utf8]{inputenc}

\usepackage{microtype}

\usepackage{inconsolata}

\usepackage{array}
\usepackage{multirow}
\usepackage{makecell}
\usepackage{soul}
\usepackage{url}
\usepackage{graphicx}
\usepackage{amsmath}
\usepackage{booktabs}
\usepackage{subfigure}
\usepackage{algorithmic}
\usepackage{enumitem}
\usepackage{amsthm,amsmath,amssymb}
\usepackage{mathrsfs}
\usepackage{bm}
\usepackage{hyperref}
\usepackage[ruled,linesnumbered]{algorithm2e}

\makeatletter
\newcommand{\thickhline}{%
    \noalign {\ifnum 0=`}\fi \hrule height 1pt
    \futurelet \reserved@a \@xhline
}

\title{Back to the Future: Bidirectional Information Decoupling Network\\ for Multi-turn Dialogue Modeling}

\author{Yiyang Li$^{1,2}$, Hai Zhao$^{1,2,}$\thanks{\; Corresponding author. This paper was partially supported by Key Projects of National Natural Science Foundation of China (U1836222 and 61733011).} \and Zhuosheng Zhang$^{1,2}$ \\
        $^1$ Department of Computer Science and Engineering, Shanghai Jiao Tong University\\
        $^2$ Key Laboratory of Shanghai Education Commission for Intelligent Interaction\\and Cognitive Engineering, Shanghai Jiao Tong University\\
        \texttt{\{eric-lee,zhangzs\}@sjtu.edu.cn,zhaohai@cs.sjtu.edu.cn}
}

\begin{document}
\maketitle

\begin{abstract}
Multi-turn dialogue modeling as a challenging branch of natural language understanding (NLU), aims to build representations for machines to understand human dialogues, which provides a solid foundation for multiple downstream tasks. Recent studies of dialogue modeling commonly employ pre-trained language models (PrLMs) to encode the dialogue history as successive tokens, which is insufficient in capturing the temporal characteristics of dialogues. Therefore, we propose Bidirectional Information Decoupling Network (BiDeN) as a universal dialogue encoder, which explicitly incorporates both the past and future contexts and can be generalized to a wide range of dialogue-related tasks. Experimental results on datasets of different downstream tasks demonstrate the universality and effectiveness of our BiDeN. The official implementation of BiDeN is available at \url{https://github.com/EricLee8/BiDeN}.
\end{abstract}

\section{Introduction}
Multi-turn dialogue modeling as one of the core tasks in natural language understanding, aims to build representations for machines to understand human dialogues. It is the foundation of solving multiple dialogue-related tasks such as selecting a response \cite{lowe2015ubuntu, zhang2018modeling, cui2020mutual}, answering questions \cite{sun2019dream, yang2019friendsqa, li2020molweni}, or making a summarization according to the dialogue history \cite{SAMSUM, chen2021dialsumm}.

Dialogue contexts possess their intrinsic nature of informal, colloquial expressions, discontinuous semantics, and strong temporal characteristics \cite{reddy2019coqa, yang2019friendsqa, chen2020neural, qin2021timedial}, making them harder for machines to understand compared to plain texts \cite{rajpurkar2016squad, cui2020mutual, zhang2021multi}. To tackle the aforementioned obstacles, most of the existing works on dialogue modeling have made efforts from three perspectives. The first group of works adopt a hierarchical encoding strategy by first encoding each utterance in a dialogue separately, then making them interact with each other by an utterance-level interaction module \cite{zhang2018modeling, li2020transformers, gu2021dialogbert}. This strategy shows sub-optimal to model multi-turn dialogue owing to the neglect of informative dialogue contexts when encoding individual utterances. The second group of works simply concatenate all the utterances chronologically as a whole (together with response candidates for the response selection task), then encode them using pre-trained language models (PrLMs) \cite{zhang-etal-2020-dialogpt, smith-etal-2020-put}. This encoding pattern has its advantage of leveraging the strong interaction ability of self-attention layer in Transformer \cite{vaswani2017attention} to obtain token-level contextualized embedding, yet ignores utterance-level modeling in dialogue contexts. \citet{sankar-etal-2019-neural} also demonstrate that the simple concatenation is likely to ignore the conversational dynamics across utterances in the dialogue history. The third group of works employ a \emph{pack} and \emph{separate} method by first encoding the whole dialogue context using PrLMs, then separating them to form representations of different granularities (turn-level, utterance-level, etc.) for further interaction \cite{zhang2021multi, liu2021filling}.

\begin{figure}[tbp]
	\includegraphics[width=0.45\textwidth]{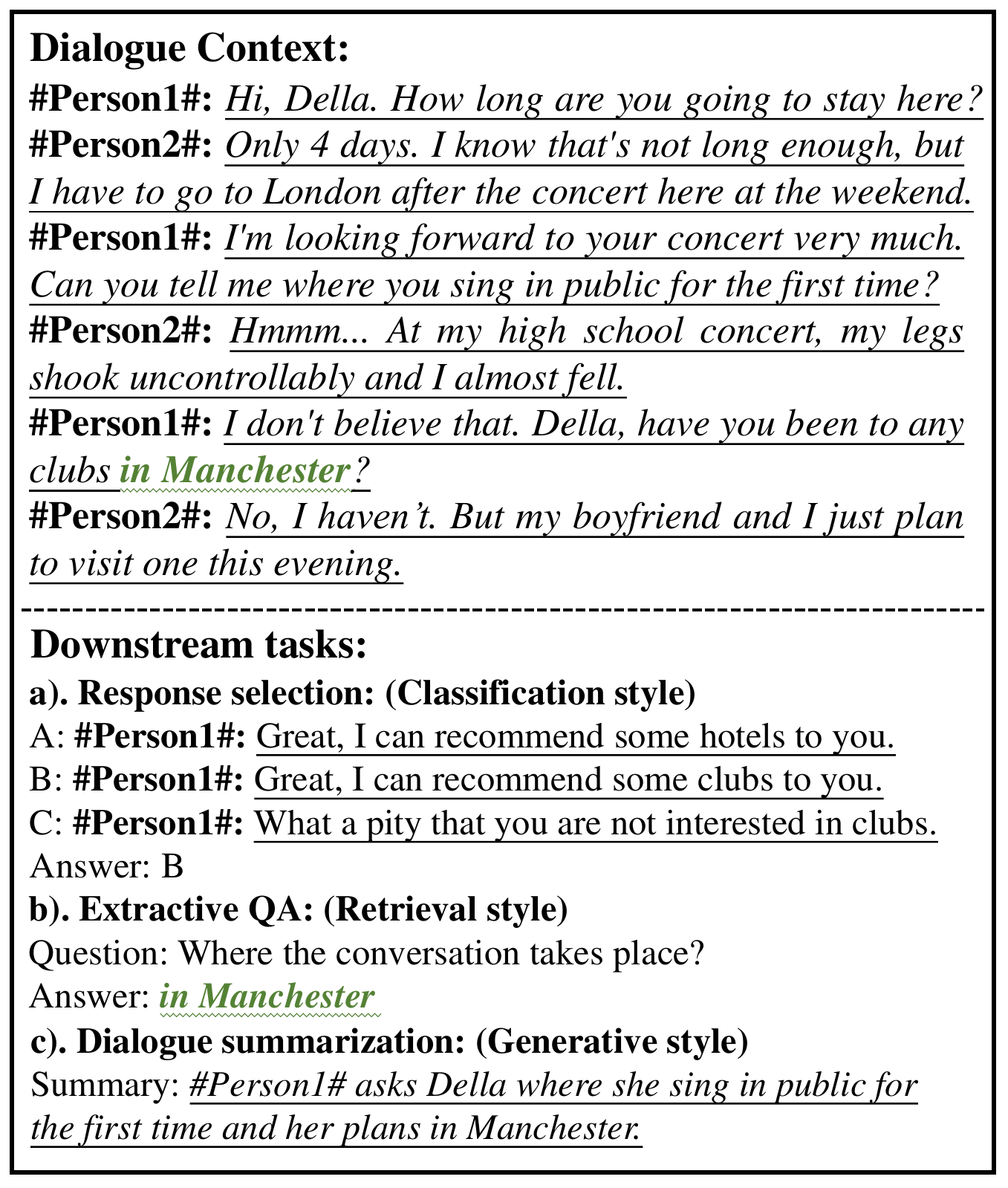}
	\centering
	\caption{An example of different downstream tasks based on dialogue contexts.} 
	\label{tasks_example}
\end{figure}

Unfortunately, all works mentioned above paid little attention to the temporal characteristics of dialogue texts, which are supposed to be useful and essential for modeling multi-turn dialogues. Different from previous works and to fill the gap of effectively capturing the temporal features in dialogue modeling, we propose a simple but effective Bidirectional Information Decoupling Network (BiDeN), which explicitly incorporates both the past and future information from the dialogue contexts. Our proposed model can serve as a universal dialogue encoder and be generalized to a wide range of downstream dialogue-related tasks covering classification, retrieval, and generative styles as illustrated in Figure \ref{tasks_example}.

In detail, we first concatenate all the utterances to form a dialogue context, then encode it with a PrLM. After obtaining the representations output by the PrLM, three additional parameter-independent information decoupling layers are applied to decouple three kinds of information entangled in the dialogue representations: past-to-current, future-to-current, and current-to-current information. Respectively, the past-to-current information guides the modeling of what the current utterance should be like given the past dialogue history, the future-to-current information guides the modeling of what kind of current utterance will lead to the development of the future dialogue, and the current-to-current information guides the modeling of the original semantic meaning resides in the current utterance. After obtaining these representations, we fuse them using a Mixture of Experts (MoE) mechanism \cite{jacobs1991adaptive} to form the final dialogue history representations.

Let's focus again on Figure \ref{tasks_example} and take the response selection task as example. When modeling the three candidate responses, the past-to-current information of the responses and the future-to-current information of each utterance in the context will detect incoherent temporal features in response \emph{A} and \emph{C}, and coherent feature of response \emph{B}, which help the model to deduce the final answer.

We conduct experiments on three datasets that belong to different types of dialogue-related tasks: Multi-Turn Dialogue Reasoning (MuTual, \citealt{cui2020mutual}) for response selection, Molweni \cite{li2020molweni} for extractive question-answering (QA) over multi-turn multi-party dialogues, and DIALOGSUM \cite{chen2021dialsumm} for dialogue summarization. Experimental results on these three datasets show that BiDeN outperforms strong baselines by large margins and achieves new state-of-the-art results.

The contributions of our work are three-fold:
\begin{itemize}[leftmargin=*, topsep=1pt]
    \setlength{\itemsep}{0pt}
    \setlength{\parsep}{0pt}
    \setlength{\parskip}{0pt}
    \item The proposed model can serve as a universal dialogue encoder and easily be applied to various downstream dialogue-related tasks.
    \item The proposed model is designed to model the indispensable temporal characteristics of dialogue contexts, which are ignored by previous works. To the best of our knowledge, this is the first paper that introduces the back-and-forth reading strategy \cite{sun-etal-2019-improving} to the modeling of temporal characteristics of dialogues.
    \item Experimental results on three benchmark datasets show that our simple but effective model outperforms strong baselines by large margins, and achieves new state-of-the-art results.
\end{itemize}

\begin{figure*}[tbp]
	\includegraphics[width=0.96\textwidth]{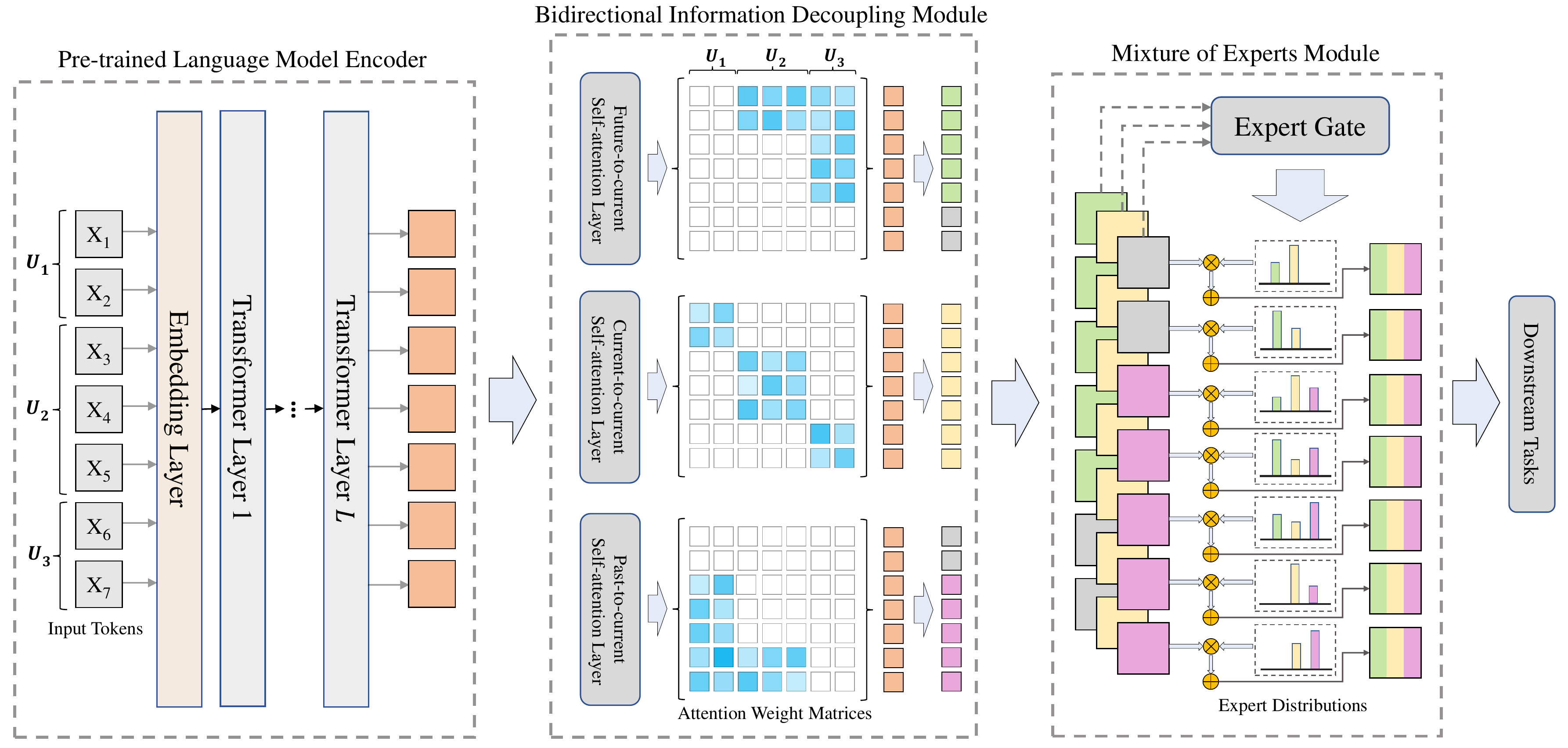}
	\centering
	\caption{The overview of our model, which consists of three main parts: a pre-trained language model encoder (PrLM encoder), a Bidirectional Information Decoupling Module (BIDM) and a Mixture of Experts (MoE) module. A gray square in the middle part means the representation of this token in this channel is invalid, which will be ignored by the MoE module.} 
	\label{overview}
\end{figure*}

\section{Related Works}
\subsection{Pre-trained Language Models}
Our model is implemented based on pre-trained language models (PrLMs), which have achieved remarkable results on many natural language understanding (NLU) tasks and are widely used as a text encoder by many researchers \cite{whq, space}. Based on self-attention mechanism and Transformer \cite{vaswani2017attention}, together with pre-training on large corpora, PrLMs have a strong capability of encoding natural language texts into contextualized representations. To name a few, BERT \cite{devlin2019bert}, ALBERT \cite{lan2019albert} and ELECTRA \cite{clark2020electra} are the most prominent ones for NLU; GPT \cite{radford2019language}, T5 \cite{raffel2020exploring} and BART \cite{lewis-etal-2020-bart} are the most representative ones for natural language generation. In our work, we select BERT, ELECTRA, and BART as the encoder backbones of our model.

\subsection{Multi-turn Dialogue Modeling}
\label{sec:mdm}
There are several previous studies on multi-turn dialogue modeling for different downstream tasks. \citet{li-etal-2021-conversations} propose DialoFlow, which utilizes three novel pre-training objectives to capture the information dynamics across dialogue utterances for response generation. \citet{zhang2021multi} design a Pivot-oriented
Deep Selection mode (PoDS) to explicitly capture salient utterances and incorporate common sense knowledge for response selection. \citet{liu2021filling} propose a Mask-based Decoupling-Fusing Network (MDFN), which adopts a mask mechanism to explicitly model speaker and utterance information for two-party dialogues. \citet{liu2021graph} propose a Graph Reasoning Network (GRN) to explicitly model the reasoning process on multi-turn dialogue response selection. Different from all these detailed works focusing on specific tasks, in this work, we devote ourselves to a universal dialogue modeling enhancement by effectively capturing the long-term ignored temporal features of dialogue data.

\section{Methodology}
In this part, we introduce BiDeN and its three modules, whose overview is shown in Figure \ref{overview}. The left part is a pre-trained language model encoder. Given a sequence of input tokens, the PrLM encoder yields their contextualized representations. The middle part is a Bidirectional Information Decoupling Module (BIDM), which decouples the entangled representations into three channels for each utterance: future-to-current representations, past-to-current representations and current-to-current representations. The right part is a Mixture of Experts (MoE) module, which calculates an expert distribution to dynamically fuse the three kinds of representations for each token. In the following sections, we will introduce them in detail, respectively.

\subsection{Pre-trained Language Model Encoder}
Given a set of input tokens $\mathbb{X} = \{w_1, w_2, ..., w_n\}$, we first embed them into a high dimensional embedding space using an embedding look-up table $\phi$: $E_T = \phi(\mathbb{X}) = \{\bm{e_1}, \bm{e_2}, ..., \bm{e_n}\} \in \mathcal{R}^{n\times d}$, where $d$ is the hidden size defined by the PrLM. After that, positional embedding $E_P$ and segment embedding $E_S$ will be added to $E_T$ to model the positional and segment information: $E = E_T + E_P + E_S$. $E$ is later fed into the Transformer layers to obtain the contextualized representations $H$. We first introduce the multi-head self-attention (MHSA) mechanism:
\begin{equation}
    \label{eq_multi}
    \begin{split}
        & \operatorname{Attn}(\bm{Q}, \bm{K}, \bm{V}) = \operatorname{softmax}(\frac{\bm{QK^T}}{\sqrt{d_k}})\bm{V}\\
        & \operatorname{head_i} = \operatorname{Attn}(E\bm{W_i^Q}, E\bm{W_i^K}, E\bm{W_i^V})\\
        & \operatorname{MultiHead}(H) = [\operatorname{head}_1,\dots,\operatorname{head}_h]\bm{W^O}
    \end{split}
\end{equation}
where $\bm{W_i^Q}\in \mathcal{R}^{d\times d_q}$, $\bm{W_i^K}\in \mathcal{R}^{d\times d_k}$, $\bm{W_i^V}\in \mathcal{R}^{d\times d_v}$, $\bm{W^O}\in \mathcal{R}^{hd_v\times d}$ are transformation matrices with trainable weights, $h$ is the number of attention heads, and $[;]$ denotes the concatenation operation. $d_q$, $d_k$, $d_v$ are the hidden sizes of the query vector, key vector and value vector, respectively. MHSA is the foundation of Transformer, which is easier to train and can model long distance dependencies. Given the input embeddings $E$, the Transformer layers $\operatorname{Trans}(E)$ is formulated as follows:
\begin{equation}
    \label{eq_transformer}
    \begin{aligned}
        & H^0 = E \in \mathcal{R}^{n\times d} \\
        & H^i_{tmp} = {\rm LN}({\rm MultiHead}(H^{i-1}) + H^{i-1})\\
        & H^i = {\rm LN}({\rm FFN}(H^i_{tmp}) + H^i_{tmp})\\
        & {\rm FFN}(x) = {\rm ReLU}(x\bm{W_1} + \bm{b_1})\bm{W_2} + \bm{b_2}
    \end{aligned}
\end{equation}
where ${\rm LN}$ is layer normalization, ${\rm ReLU}$ is a non-linear activation function and $\bm{W_1}$, $\bm{W_2}$, $\bm{b_1}$, $\bm{b_2}$ are trainable linear transformation matrices and bias vectors, respectively.

We denote the stack of $L$ Transformer layers as $\operatorname{Trans-L}$, the final representation output by the PrLM encoder is:
\begin{equation}
    H = \operatorname{Trans-L}(E) \in \mathcal{R}^{n\times d}
\end{equation}

\subsection{Bidirectional Information Decoupling}
\label{BIDM}
Given the token representations output by the PrLM encoder, the Bidirectional Information Decoupling Module will decouple them into three channels in a back-and-forth way. We first introduce a masked Transformer layer $\operatorname{MTrans}(E, M)$ by modifying the first equation on Eq. (\ref{eq_multi}) to:
\begin{equation}
    \operatorname{Attn}(\bm{Q}, \bm{K}, \bm{V}) = \operatorname{softmax}(\frac{\bm{QK^T}}{\sqrt{d_k}}+M)\bm{V}
\end{equation}
where $M$ is an $n\times n$ attention mask matrix. 
The function of $M$ is to convert the original fully-connected attention graphs to partially-connected ones, so that each token will be forced to only focus on part of the input sequence. Here we introduce three kinds of attention masks, which guide the decoupling process of the future-to-current channel, current-to-current channel, and past-to-current channel, respectively. Specifically, suppose $I(i)$ means the index of the utterance that the $i_{th}$ token belongs to, the three kinds of masks are obtained by:
\begin{equation}
    \label{eq_masks}
    \begin{aligned}
        & M_{f2c}[i, j]=\left\{
        \begin{array}{cl}
            0, & \text {if}\ I(i) < I(j) \\
            -\infty, & \text {otherwise}
        \end{array}\right. \\
        & M_{c2c}[i, j]=\left\{
        \begin{array}{cl}
            0, & \text {if}\ I(i) = I(j) \\
            -\infty, & \text {otherwise}
        \end{array}\right. \\
        & M_{p2c}[i, j]=\left\{
        \begin{array}{cl}
            0, & \text {if}\ I(i) > I(j) \\
            -\infty, & \text {otherwise}
        \end{array}\right.
    \end{aligned}
\end{equation}
where $M_{f2c}$, $M_{c2c}$ and $M_{p2c}$ are future-to-current mask, current-to-current mask and past-to-current mask, respectively. After obtaining these masks, three parameter-independent $\operatorname{MTrans-1}(H, M)$ are applied to decouple the original representation $H$ as follows:
\begin{equation}
    \begin{aligned}
        & H_{f2c} = \operatorname{MTrans-1^{f2c}}(H, M_{f2c}) \\
        & H_{c2c} = \operatorname{MTrans-1^{c2c}}(H, M_{c2c}) \\
        & H_{p2c} = \operatorname{MTrans-1^{p2c}}(H, M_{p2c})
    \end{aligned}
\end{equation}

Note that there are tokens who has no connections to any tokens in certain channels, e.g. the tokens of the first utterance has no connections to other tokens in past-to-future channel since there are no previous utterances. To handle this case, we simply ignore the invalid representations (gray squares in Figure \ref{overview}) by adding a fusion mask during the fusion process, which will be introduced in Section \ref{moe}.

After the decoupling process, $H_{p2c}$ contains the information of the influence that the past dialogue history brings about to the current utterance, or in other words, it reflects what the current utterance should be like given the past dialogue history. $H_{f2c}$ contains the information of the influence that the current utterance brings about to future dialogue contexts, or put it another way, it reflects what kind of current utterance will lead to the development of the future dialogue. Finally, $H_{c2c}$ contains the information of the original semantic meaning resides in the current utterance. By explicitly incorporating past and future information into each utterance, our BIDM is equipped with the ability to capture temporal features in dialogue contexts.

\subsection{Mixture of Experts Module}
\label{moe}
We first introduce the Mixture of Experts (MoE) proposed by  \citet{jacobs1991adaptive}. Specifically, $m$ experts $\{f_i(x)\}_{i=1}^m$ are learned to handle different input cases. Then a gating function $G = \{g_i(x)\}_{i=1}^m$ are applied to determine the importance of each expert dynamically by assigning weights to them. The final output of MoE is the linear combination of each expert:
\begin{equation}
    \label{eq_moe}
    MoE(x) = \sum_{i=1}^m\ g_i(x)\cdot f_i(x)
\end{equation}

In this work, $\operatorname{MTrans^{f2c}}$, $\operatorname{MTrans^{c2c}}$ and $\operatorname{MTrans^{p2c}}$ are treated as three experts. We design the gating function similar as \citet{liu2021filling} that utilizes the original output $H$ to guide the calculation of expert weights. In detail, we first calculate a heuristic matching representation between $H$ and the three outputs of Section \ref{BIDM}, respectively, then obtain the expert weights $G$ by considering all three matching representations and calculate the final fused representation $H_e$ as follows:
\begin{equation}
    \label{eq_experts}
    \begin{aligned}
        & \operatorname{Heuristic}(X, Y) = [X; Y; X-Y; X\odot Y]\\
        & S_{f} = \operatorname{ReLU}(\operatorname{Heuristic}(H, H_{f2c}) \bm{W_f} + \bm{b_f}) \\
        & S_{c} = \operatorname{ReLU}(\operatorname{Heuristic}(H, H_{c2c}) \bm{W_c} + \bm{b_c}) \\
        & S_{p} = \operatorname{ReLU}(\operatorname{Heuristic}(H, H_{p2c}) \bm{W_p} + \bm{b_p}) \\
        & G = \operatorname{Softmax}([S_{f}; S_{c}; S_{p}] \bm{W_g} + M_g) \in \mathcal{R}^{n\times d\times 3}\\
        & H_e = \operatorname{Sum}(\operatorname{Stack}(H_{f2c};H_{c2c};H_{p2c})\odot G)
    \end{aligned}
\end{equation}

Here $H_e\in \mathcal{R}^{n\times d}$, $\odot$ represents element-wise multiplication, $\bm{W_f}$, $\bm{W_c}$, $\bm{W_p} \in \mathcal{R}^{4d\times d}$ and $\bm{b_f}$, $\bm{b_c}$, $\bm{b_p}\in \mathcal{R}^d$ are trainable transformation matrices and bias vectors, respectively. $\bm{W_g}\in \mathcal{R}^{3d\times d\times 3}$ is a trainable gating matrix that generates feature-wise expert scores by considering all three kinds of information. $M_g$ is a fusion mask added for ignoring invalid tokens, which is introduced in Section \ref{BIDM}.

%============================================
% For classification tasks (e.g. response selection), we adopt Eq. (\ref{eq_experts}) to fuse the representations. For retrieval and generative tasks, which focus more on token-level information, we calculate the token-wise expert weights instead of the feature-wise ones by modifying the last two equations of Eq. (\ref{eq_experts}) to:
% \begin{equation}
%     \label{eq_tokenwise}
%     \begin{aligned}
%         & G_t = \operatorname{Softmax}([S_{f}; S_{c}; S_{p}] W_{gt}+M_{gt}) \in \mathcal{R}^{n\times 3}\\
%         & H_e = \operatorname{Sum}(\operatorname{Stack}(H_{f2c};H_{c2c};H_{p2c})\odot \operatorname{Expand}(G_t))
%     \end{aligned}
% \end{equation}
% where $W_{gt}\in \mathcal{R}^{3d\times 3}$ is a trainable token-level gating matrix.
%============================================

After incorporating future-to-current, past-to-current and current-to-current information, we obtain temporal-aware representation $H_e$, which can be used for various dialogue-related tasks described in Section \ref{experiments}.

\section{Experiments}
\label{experiments}

\begin{table*}[tbp]
    \centering
    \small
    \begin{tabular}{lrrrrrrr}
        \specialrule{0.09em}{0.0pt}{1.8pt}
        \multirow{2}{*}{Model} & \multicolumn{3}{c} {\textbf{MuTual}} & & \multicolumn{3}{c} {\textbf{MuTual}$\operatorname{^{plus}}$} \\
        & \textbf{R@1} & \textbf{R@2} & \textbf{MRR} & & \textbf{R@1} & \textbf{R@2} & \textbf{MRR} \\
        \specialrule{0.03em}{1.3pt}{0.5pt}
        \specialrule{0.03em}{0.5pt}{1.3pt}
        \emph{From Paper} \cite{cui2020mutual} & & & & & & \\
        DAM & 0.239 & 0.463 & 0.575 & & 0.261 & 0.520 & 0.645 \\
        SMN & 0.274 & 0.524 & 0.575 & & 0.264 & 0.524 & 0.578 \\
        BERT & 0.657 & 0.867 & 0.803 & & 0.514 & 0.787 & 0.715 \\
        RoBERTa & 0.695 & 0.878 & 0.824 & & 0.622 & 0.853 & 0.782 \\
        \specialrule{0.06em}{1.3pt}{1.3pt}
        % \emph{Our Implementation}\\
        ELECTRA & 0.907 & 0.975 & 0.949 & & 0.826 & 0.947 & 0.904 \\
        \quad +BIDM & $^*$0.916 & $^*$\textbf{0.980} & $^*$0.955 & & 0.830 & 0.950 & 0.906\\
        \quad \quad +BiDeN & $^{**}$\textbf{0.935} & $^*$0.979 & $^{**}$\textbf{0.963} & & $^{**}$\textbf{0.839} & \textbf{0.951} & $^*$\textbf{0.910} \\
        \specialrule{0.09em}{1.2pt}{0.0pt}
    \end{tabular}
    \caption{Results on the development sets of MuTual and MuTual$\operatorname{^{plus}}$. The first four rows are directly taken from the original paper of MuTual. Here $^*$ denotes that the result outperforms the baseline model significantly with \emph{p-value} $<0.05$ in paired t-test and $^{**}$ denotes $<0.01$.}
    \label{tab_mutual_dev}
\end{table*}

\subsection{Benchmark Datasets}
We adopt Multi-Turn Dialogue Reasoning (Mutual, \citealt{cui2020mutual}) for response selection, Molweni \cite{li2020molweni} for extractive QA over multi-turn multi-party dialogues, and DIALOGSUM \cite{chen2021dialsumm} for dialogue summarization.

\textbf{MuTual} is proposed to boost the research of the reasoning process in retrieval-based dialogue systems. It consists of 8,860 manually annotated two-party dialogues based on Chinese student English listening comprehension exams. For each dialogue, four response candidates are provided and only one of them is correct. A \emph{plus} version of this dataset is also annotated by randomly replacing a candidate response with \emph{safe response} (e.g. \emph{I didn't hear you clearly}), in order to test whether a model is able to select a safe response when the other candidates are all inappropriate. This dataset is more challenging than other datasets for response selection since it requires some reasoning to select the correct candidate. This is why we choose it as our benchmark for the response selection task.

\textbf{Molweni} is a dataset for extractive QA over multi-party dialogues. It is derived from the large-scale multi-party dialogue dataset --- Ubuntu Chat Corpus \cite{lowe2015ubuntu}, whose main theme is technical discussions about problems on the Ubuntu system. In total, it contains 10,000 dialogues annotated with questions and answers. Given a dialogue, several questions will be asked and the answer is guaranteed to be a continuous span in the dialogue context. The reason we choose this dataset as a benchmark for retrieval style task is that we want to test whether our model still holds on multi-party dialogue contexts.

\textbf{DIALOGSUM} is a large-scale real-life dialogue summarization dataset. It contains 13,460 daily conversations collected from different datasets or websites. For each dialogue context, annotators are asked to write a concise summary that conveys the most salient information of the dialogue from an observer's perspective. This dataset is designed to be highly abstractive, which means a generative model should be adopted to generate the summaries.

\subsection{Experimental Setups}
On the MuTual dataset, ELECTRA is adopted as the PrLM encoder for a fair comparison with previous works. We follow \citet{liu2021filling} to get the dialogue-level representation $H_d$ from $H_e$. We first obtain the utterance-level representations by applying a max-pooling over the tokens of each utterance, then use a Bidirectional Gated Recurrent Unit (Bi-GRU) to summarize the utterance-level representations into a single dialogue-level vector. For one dialogue history with four candidate responses, we concatenate them to form four dialogue contexts and encode them to obtain $H_D = \{H_d^i\}_{i=1}^4 \in \mathcal{R}^{d\times 4}$. Given the index of the correct answer $i^{target}$, we compute the candidate distribution and classification loss by:
\begin{equation}
    \begin{aligned}
        & P_D = \operatorname{Softmax}(\bm{w_d}^T H_D) \in \mathcal{R}^4\\
        & \mathcal{L}_D = -log(P_D[i^{target}])
    \end{aligned}
\end{equation}
where $\bm{w_d}\in \mathcal{R}^{d}$ is a trainable linear classifier and $\mathcal{L}_D$ is the cross entropy loss.

On the Molweni dataset, BERT is adopted as the PrLM encoder for a fair comparison with previous works. We simply regard the question text as a special utterance and concatenate it to the end of the dialogue history to form the input sequence. After obtaining $H_e$, we add two linear classifiers to compute the start and end distributions over all tokens. Given the start and end positions of the answer span $[a_s, a_e]$, cross entropy loss is adopted to train our model:
\begin{equation}
    \begin{aligned}
        & P_{start} = \operatorname{Softmax}(H_e \bm{w_s}^T) \in \mathcal{R}^n \\
        & P_{end} = \operatorname{Softmax}(H_e \bm{w_e}^T) \in \mathcal{R}^n \\
        & \mathcal{L}_{SE} = -(\text{log}(P_{start}[a_s])+\text{log}(P_{end}[a_e]))
    \end{aligned}
\end{equation}
where $\bm{w_s}$ and $\bm{w_e}\in \mathcal{R}^d$ are two trainable linear classifiers.

On the DIALOGSUM dataset, BART is chosen as our backbone since it is one of the strongest generative PrLMs. Different from the previous two PrLMs, BART adopts an encoder-decoder architecture where the encoder is in charge of encoding the input texts and the decoder is responsible for generating outputs. Therefore, we add our BIDM after the encoder of BART. Note that BART is pre-trained on large corpora using self-supervised text denoising tasks, hence there is a strong coupling on the pre-trained parameter weights between the encoder and decoder. Under this circumstance, simply adding our BIDM after the encoder will destroy the coupling between encoder and decoder, resulting in the decline of model performance. To tackle this problem, we propose novel a copy-and-reuse way to maintain the parameter-wise coupling between the encoder and decoder. Specifically, instead of using randomly initialized decoupling layers, we reuse the last layer of BART encoder and load the corresponding pre-trained weights to initialize the future-to-current, current-to-current, and past-to-current decoupling layers, respectively.
We train this model by an auto-regressive language model loss:
\begin{equation}
    \mathcal{L}_G= -\sum_{t=1}^{N} \log p\left(w_{t} \mid \bm{\theta}, w_{<t}\right)
\end{equation}
where $\bm{\theta}$ is the model parameters, $N$ is the total number of words in the target summary and $w_t$ is the token at time step $t$. We also conduct experiments on the SAMSum \cite{SAMSUM} dataset, and the results are presented in Appendix \ref{app:samsum}.

For hyper-parameter settings and more details about our experiments, please refer to Appendix \ref{app:hyper}.

\subsection{Results}
In this section, we will briefly introduce the baseline models and evaluation metrics, then present the experimental results on different datasets.

\subsubsection{Results on MuTual}
Table \ref{tab_mutual_dev} shows the results on the development sets of MuTual and MuTual$\operatorname{^{plus}}$, respectively. Following \citet{cui2020mutual}, we adopt \textbf{R@k} (Recall at K) and \textbf{MRR} (Mean Reciprocal Rank) as our evaluation metrics. The baseline models we compare here are: two PrLM-free methods DAM \cite{dam} and Sequential Matching Network (SMN, \citealt{wu2017sequential}), who encode the context and response separately and match them on different granularities. Three PrLM-based baselines: BERT, RoBERTa \cite{liu2019roberta} and ELECTRA. We see from Table \ref{tab_mutual_dev} that PrLM-free models perform worse than PrLM-based models and different PrLMs have different results, where ELECTRA is the best. Compared with vanilla ELECTRA, simply adding BIDM is able to improve the performance, demonstrating that explicitly incorporating the temporal features has a heavy impact on understanding dialogue contexts. By further equipping BiDeN, we observe giant improvements over ELECTRA by $\mathbf{2.8}$\textbf{\%} and $\mathbf{1.3}$\textbf{\%} R@1 on MuTual and MuTual$\operatorname{^{plus}}$, respectively. Note that the absolute improvements on R@2 are not as high as on R@1. We infer this is because the scores on this metric are already high enough, thus it is harder to achieve very large absolute improvements. However, when it comes to the error rate reduction, BiDeN impressively reduces the error rate from 2.5\% to 2.0\%, which is a 20\% relative reduction.

\begin{table}[tbp]
    \centering
    \small
    \begin{tabular}{l c c c}
        \specialrule{0.09em}{0.0pt}{1.8pt}
        \multirow{2}{*}{Model} & \multicolumn{3}{c} {\textbf{MuTual / \textbf{MuTual}$\operatorname{^{plus}}$}} \\
        & \textbf{R@1} & \textbf{R@2} & \textbf{MRR}\\
        \specialrule{0.03em}{1.3pt}{0.5pt}
        \specialrule{0.03em}{0.5pt}{1.3pt}
        GRN & 0.915 / 0.841 & 0.983 / 0.957 & 0.954 / 0.913\\
        MDFN & 0.916 / --- & 0.984 / --- & 0.956 / ---\\
        DAPO & 0.916 / 0.836 & \textbf{0.988} / 0.955 & 0.956 / 0.910\\
        CF-DR & 0.921 / 0.810 & 0.985 / 0.946 & 0.958 / 0.896\\
        \textbf{BiDeN} & \textbf{0.930} / \textbf{0.845} & 0.983 / \textbf{0.958} & \textbf{0.962} / \textbf{0.914}\\
        \specialrule{0.09em}{1.2pt}{0.0pt}
    \end{tabular}
    \caption{Results on the hidden test sets from the \emph{leaderboard} of MuTual dataset.}
    \label{tab_mutual_test}
\end{table}

Table \ref{tab_mutual_test} presents the current SOTA models on the leaderboard of MuTual, which is tested on the hidden test set. Graph Reasoning Network (GRN, \citealt{liu2021graph}) utilizes Graph Convolutional Networks to model the reasoning process. MDFN \cite{liu2021filling} is introduced in Section \ref{sec:mdm}, Dialogue-Adaptive Pre-training Objective (DAPO, \citealt{li2020task}) designs a special pre-training objective for dialogue modeling. CF-DR is the previous first place on the leaderboard, but without a publicly available paper. We see from the table that BiDeN achieves new SOTA results on both datasets, especially on MuTual, where we observe a performance gain of \textbf{0.9\%} R@1 score.

\subsubsection{Results on Molweni}
Table \ref{tab_molweni} shows the results on Molweni dataset, where we use \textbf{Exactly Match (EM)} and \textbf{F1} score as the evaluation metrics. DADGraph \cite{li2021dadgraph} utilizes the discourse parsing annotations in the Molweni dataset and adopts Graph Neural Networks (GNNs) to explicitly model the discourse structure of dialogues. Compared with them, BiDeN needs no additional discourse labels but performs better. SUP \cite{sup} designs auxiliary self-supervised predictions of speakers and utterances to enhance multi-party dialogue comprehension. We see from the table that our model outperforms vanilla BERT by large margins, which are \textbf{2.2\%} on EM and \textbf{2.5\%} on F1, respectively. In addition, SUP can be further enhanced by BiDeN.

\begin{table}[tbp]
    \centering
    \small
    \begin{tabular}{l r r}
        \specialrule{0.09em}{0.0pt}{2.2pt}
        Model & \textbf{EM} & \textbf{F1}\\
        \specialrule{0.03em}{1.3pt}{0.5pt}
        \specialrule{0.03em}{0.5pt}{1.3pt}
        BERT$_{\operatorname{DADGraph}}$ \cite{li2021dadgraph} & 0.465 & 0.615\\
        BERT$_{\operatorname{SUP}}$ \cite{sup} & 0.492 & 0.640 \\
        % BERT$_{\operatorname{R-GCN}}$ \cite{maxb} & 0.497 & 0.644 \\
        BERT & 0.458 & 0.602\\
        \quad +BIDM & $^*$0.475 & $^{**}$0.626\\
        \quad \quad +BiDeN & $^{**}$0.481 & $^{**}$0.632\\
        SUP+BiDeN & $^{**}$\textbf{0.503} & $^{**}$\textbf{0.659} \\
        \specialrule{0.09em}{1.2pt}{0.0pt}
    \end{tabular}
    \caption{Results on Molweni, where $^*$ and $^{**}$ represent the same as in Table \ref{tab_mutual_dev}.}
    \label{tab_molweni}
\end{table}

\begin{table}[tbp]
    \centering
    \small
    \begin{tabular}{l r r r}
        \specialrule{0.09em}{0.0pt}{2.0pt}
        Model & \textbf{Rouge-1} & \textbf{Rouge-2} & \textbf{Rouge-L}\\
        \specialrule{0.03em}{1.3pt}{0.5pt}
        \specialrule{0.03em}{0.5pt}{1.3pt}
        DialoBART & 0.533 & 0.296 & 0.520\\
        DialSent-PGG & 0.547 & 0.305 & \textbf{0.535}\\
        BART & 0.528 & 0.289 & 0.511 \\
        \quad +BIDM & 0.535 & $^*$0.301 & $^{*}$0.523\\
        \quad \quad +BiDeN & $^{**}$\textbf{0.548} & $^{**}$\textbf{0.307} & $^{**}$0.532\\
        \specialrule{0.09em}{1.2pt}{0.0pt}
    \end{tabular}
    \caption{Results on DIALOGSUM, where $^*$ and $^{**}$ represent the same as in Table \ref{tab_mutual_dev}.}
    \label{tab_dialogsum}
\end{table}

\subsubsection{Results on DIALOGSUM}
\label{sec:result_dialogsum}
Table \ref{tab_dialogsum} presents the results on DIALOGSUM. We follow \citet{chen2021dialsumm} to adopt \textbf{Rouge} (py-rouge) as our evaluation metric, which is widely used in dialogue summarization field \cite{SAMSUM,chen2021dialsumm}.
Rouge-n computes the overlapping ratio of n-grams between the prediction and reference summaries. ROUGE-L computes the longest common subsequence (LCS) between the candidates and references, then calculates the F1 ratio by measuring the recall over references and precision over candidates. Following \cite{DiaSent}, we compute the maximum Rouge score among all references for each sample.
% Note that different from the original paper of this dataset \cite{chen2021dialsumm}, when multiple (three) references are given, we compute the maximum scores over all references rather than average them. We argue this is more reasonable since the pair-wise Rouge scores of the three references are low but all references make sense according to the paper. Simply averaging the Rouge scores is unfair to a good summary whose usage of language is different from the other two references.
Table \ref{tab_dialogsum} shows our model again outperforms the strong baseline BART by large margins, with over \textbf{2.0\%} improvements on all metrics. Besides, compared with the current SOTA models, BiDeN also exhibits its superior capability in summarizing dialogue texts. DialoBART \cite{dialogBart} utilizes DialoGPT \cite{DIALOGPT} to annotate keywords, redundant utterances and topic transitions in a dialogue, then explicitly incorporates them into the dialogue texts to train BART. Their work requires annotators to extract additional knowledge, while our BiDeN still outperforms it on all metrics. DialSent-PGG \cite{DiaSent} designs a pseudo-paraphrasing process to generate more dialogue-summary pairs from the original dataset, then post-trains the model on the pseudo-summary dataset. After post-training, they fine-tune the summarization model on the original dataset. Compared with their work, which requires an additional post-training process, BiDeN is much simpler and faster to train, yet achieves comparable results.

\section{Analysis}
In this section, we conduct experiments on MuTual dataset to get an in-depth understanding of BiDeN.

\subsection{Ablation Study}
\begin{table}[t]
    \centering
    \small
    \begin{tabular}{l c c c}
        \specialrule{0.09em}{0.0pt}{2.0pt}
        Model & \textbf{R@1} & \textbf{R@2} & \textbf{MRR}\\
        \specialrule{0.03em}{1.3pt}{0.5pt}
        \specialrule{0.03em}{0.5pt}{1.3pt}
        BiDeN & \textbf{0.935} & 0.979 & \textbf{0.963} \\
        \quad w/o BIDM & 0.912 & 0.976 & 0.953\\
        \quad w/o BIDM (same \#params) & 0.923 & \textbf{0.981} & 0.958\\
        \quad w/o MoE & 0.916 & 0.980 & 0.955\\
        \quad w/o Bi-GRU & 0.927 & 0.980 & 0.960\\
        \specialrule{0.09em}{1.2pt}{0.0pt}
    \end{tabular}
    \caption{Ablation study on development set of MuTual}
    \label{tab_ablation}
\end{table}

\begin{table}[tbp]
    \centering
    \small
    \begin{tabular}{l c c c}
        \specialrule{0.09em}{0.0pt}{2.0pt}
        Model & \textbf{R@1} & \textbf{R@2} & \textbf{MRR}\\
        \specialrule{0.03em}{1.3pt}{0.5pt}
        \specialrule{0.03em}{0.5pt}{1.3pt}
        ELECTRA & 0.907 & 0.975 & 0.949 \\
        \quad + Bi-LSTM & 0.912 & 0.977 & 0.952 \\
        \quad + Bi-GRU & 0.915 & 0.978 & 0.955 \\
        \quad + BiDeN & \textbf{0.935} & \textbf{0.979} & \textbf{0.963} \\
        \specialrule{0.09em}{1.2pt}{0.0pt}
    \end{tabular}
    \caption{Results of naive temporal modeling}
    \label{tab_naive}
\end{table}

To investigate the effectiveness of temporal modeling, we remove BIDM to see how it affects the performance. A sharp performance drop of \textbf{$2.3\%$} is observed on R@1, demonstrating the necessity and significance of explicit temporal modeling. In order to probe into whether the performance gain comes from the increment of model parameters, we conduct experiments by simply replacing the three kinds of masks defined in Eq. (\ref{eq_masks}) with all-zero masks (fully-connected attention graphs). We see from the table that the increment of parameters does add to the performance. Nevertheless, it is sub-optimal compared with explicitly modeling the temporal features by our BIDM.

We also remove MoE to see whether the dynamic fusion mechanism helps. Specifically, we replace this module with a simple mean pooling over the three decoupled representations. Result shows that MoE makes a huge contribution to the final result. To explore the effect that the task-specific design, Bi-GRU, brings about to our model, we remove the Bi-GRU and simply average the utterance representations to get the dialogue-level vector. We see from the table that Bi-GRU does have positive effects on the final performance, yet only to a slight extent compared with other modules.

\begin{figure}[tbp]
	\includegraphics[width=0.48\textwidth]{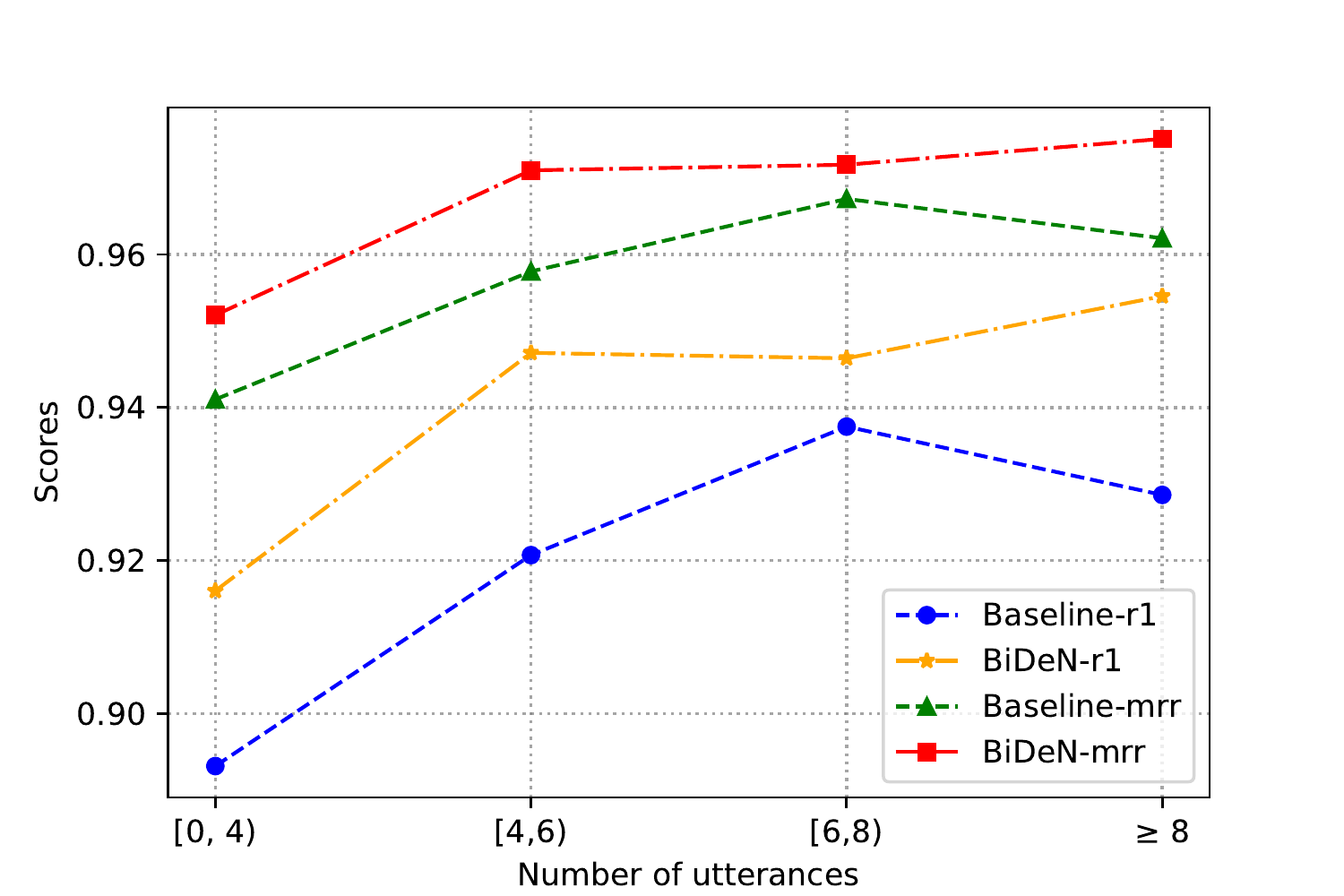}
	\centering
	\caption{Model performance v.s. the number of utterances in a dialogue, where the post-fix -r1 represents the \textbf{R@1} score and -mrr stands for the \textbf{MRR} score.} 
	\label{fig_utternum}
\end{figure}

\subsection{Naive Temporal Modeling}
When it comes to bidirectional temporal modeling, the simplest way is to use Bidirectional Recurrent Neural Networks (Bi-RNNs). To investigate whether BiDeN can be replaced by these naive temporal modeling methods, we conduct experiments by adding Bi-LSTM or Bi-GRU on top of PrLMs instead of BiDeN. We see from Table \ref{tab_naive} that utilizing Bi-RNNs can improve the performance slightly, but they are far behind BiDeN. This is because RNNs model the bidirectional information only at token-level, while BiDeN models them by explicitly modeling the utterance boundary with attention masks, which is more consistent with the data characteristics of dialogue texts.

\subsection{Influence of Dialogue Length}
Intuitively, with longer dialogue contexts comes more complicated temporal features. Based on this point, we analyze the model performance with regard to the number of utterances in a dialogue. As illustrated in Figure \ref{fig_utternum}, the scores first increase from short dialogues to medium-length dialogues. This is because medium-length dialogues contain more information for response matching than short ones. For long dialogues, the baseline model suffers a huge performance drop (see the blue and green lines), while our BiDeN keeps bringing performance improvement, demonstrating a strong ability of it to capture complicated temporal features.

\subsection{Visualization of Attentions}
\begin{figure}[tbp]
	\centering
	\subfigure[Current-to-past Attention]{
		\begin{minipage}{0.47\textwidth} 
            \includegraphics[width=0.96\textwidth]{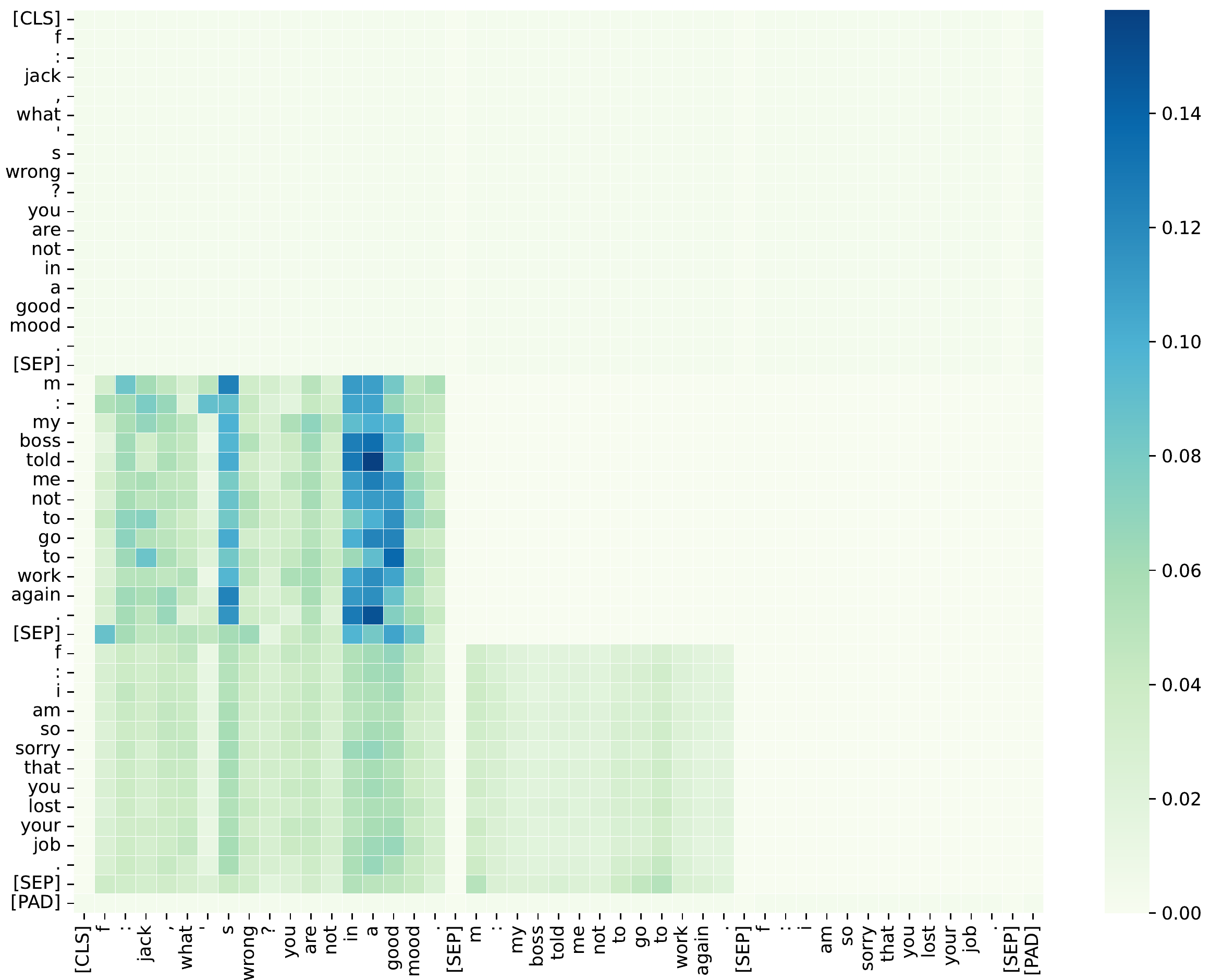} \\
		\end{minipage}
	}

	\subfigure[Current-to-future Attention]{
		\begin{minipage}{0.47\textwidth} 
            \includegraphics[width=0.96\textwidth]{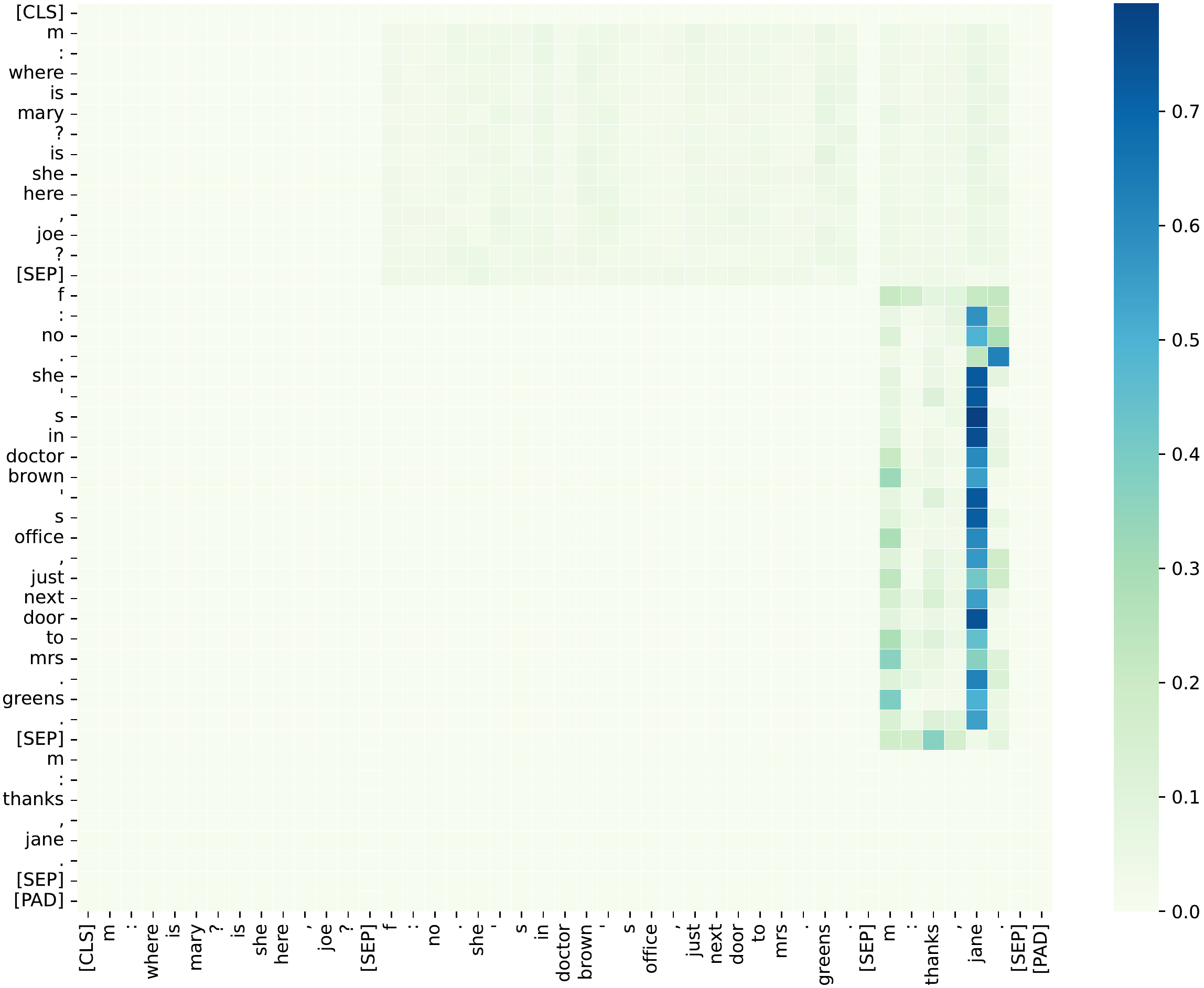} \\
		\end{minipage}
	}
	\caption{Visualization of attention weights.}
	\label{fig_attnvis}
\end{figure}

To intuitively investigate how BiDeN works, we visualize the attention weights of both current-to-past and current-to-future attentions. Figure \ref{fig_attnvis} (a) shows the current-to-past attention weights. We see that the utterance \emph{My boss told me not to go to work again} tends to focus on \emph{not in a good mood} of the previous utterance, which is a causal discovery. Similarly, the last utterance \emph{I am so sorry that you lost your job} focuses more on \emph{not in a good mood} and \emph{not to go to work}. Figure \ref{fig_attnvis} (b) shows an example of current-to-future attention, which is an incorrect response example taken from MuTual dataset. We see that the current utterance pays great attention on the name \emph{Jane}, which is supposed to be \emph{Joe}. This observation indicates that BiDeN is capable of detecting the logical errors in the future responses that contradict previous utterances. For more visualizations, please refer to Appendix \ref{app:more_vis}.

\section{Conclusion}
In this paper, we propose Bidirectional Information Decoupling Network (BiDeN) to explicitly model the indispensable temporal characteristics of multi-turn dialogues, which have been ignored for a long time by existing works. BiDeN shows simple but effective to serve as a universal dialogue encoder for a wide range of dialogue-related tasks. Experimental results and comprehensive analyses on several benchmark datasets have justified the effectiveness of our model.

\section*{Limitations}
Despite the contributions of our work, there are also unavoidable limitations of it.

First, we claim our BiDeN as a universal dialogue encoder which can be used in multiple dialogue-related tasks. In our paper, without the loss of generality, we select three most representative tasks in classification style, retrieval style, and generative style tasks, respectively. However, there are still so many other tasks such as dialogue emotion recognition and dialogue act classification \cite{cogat}, and also so many other large-scale datasets such as Ubuntu, Douban or E-Commerce \cite{lowe2015ubuntu, zhang2018modeling, wu2017sequential}. Due to the lack of computational resources and page limits, our BiDeN is not tested on them. We leave them to the readers who are interested in our model and encourage them to utilize our BiDeN in these tasks.

Second, the three decoupling layers and the MoE gates add to additional number of parameters (from 348M to 408M), resulting in the increment of computational overheads during training and inference (1.2$\times$ slower, 1.2$\times$ of GPU memory consumption). However, we argue that the performance gains are worth the additional overheads.

\clearpage
\bibliography{anthology,custom}

\begin{thebibliography}{43}
\expandafter\ifx\csname natexlab\endcsname\relax\def\natexlab#1{#1}\fi

\bibitem[{Chen and Yang(2020)}]{multi-view}
Jiaao Chen and Diyi Yang. 2020.
\newblock \href {https://doi.org/10.18653/v1/2020.emnlp-main.336} {Multi-view
  sequence-to-sequence models with conversational structure for abstractive
  dialogue summarization}.
\newblock In \emph{Proceedings of the 2020 Conference on Empirical Methods in
  Natural Language Processing (EMNLP)}, pages 4106--4118, Online. Association
  for Computational Linguistics.

\bibitem[{Chen et~al.(2020)Chen, Zhang, Mao, and Xu}]{chen2020neural}
Junfan Chen, Richong Zhang, Yongyi Mao, and Jie Xu. 2020.
\newblock \href {https://doi.org/10.18653/v1/2020.findings-emnlp.142} {Neural
  dialogue state tracking with temporally expressive networks}.
\newblock In \emph{Findings of the Association for Computational Linguistics:
  EMNLP 2020}, pages 1570--1579, Online. Association for Computational
  Linguistics.

\bibitem[{Chen et~al.(2021)Chen, Liu, Chen, and Zhang}]{chen2021dialsumm}
Yulong Chen, Yang Liu, Liang Chen, and Yue Zhang. 2021.
\newblock \href {https://doi.org/10.18653/v1/2021.findings-acl.449}
  {{D}ialog{S}um: {A} real-life scenario dialogue summarization dataset}.
\newblock In \emph{Findings of the Association for Computational Linguistics:
  ACL-IJCNLP 2021}, pages 5062--5074, Online. Association for Computational
  Linguistics.

\bibitem[{Clark et~al.(2020)Clark, Luong, Le, and Manning}]{clark2020electra}
Kevin Clark, Minh{-}Thang Luong, Quoc~V. Le, and Christopher~D. Manning. 2020.
\newblock \href {https://openreview.net/forum?id=r1xMH1BtvB} {{ELECTRA:}
  pre-training text encoders as discriminators rather than generators}.
\newblock In \emph{8th International Conference on Learning Representations,
  {ICLR} 2020, Addis Ababa, Ethiopia, April 26-30, 2020}. OpenReview.net.

\bibitem[{Cui et~al.(2020)Cui, Wu, Liu, Zhang, and Zhou}]{cui2020mutual}
Leyang Cui, Yu~Wu, Shujie Liu, Yue Zhang, and Ming Zhou. 2020.
\newblock \href {https://doi.org/10.18653/v1/2020.acl-main.130} {{M}u{T}ual: A
  dataset for multi-turn dialogue reasoning}.
\newblock In \emph{Proceedings of the 58th Annual Meeting of the Association
  for Computational Linguistics}, pages 1406--1416, Online. Association for
  Computational Linguistics.

\bibitem[{Devlin et~al.(2019)Devlin, Chang, Lee, and
  Toutanova}]{devlin2019bert}
Jacob Devlin, Ming-Wei Chang, Kenton Lee, and Kristina Toutanova. 2019.
\newblock \href {https://doi.org/10.18653/v1/N19-1423} {{BERT}: Pre-training of
  deep bidirectional transformers for language understanding}.
\newblock In \emph{Proceedings of the 2019 Conference of the North {A}merican
  Chapter of the Association for Computational Linguistics: Human Language
  Technologies, Volume 1 (Long and Short Papers)}, pages 4171--4186,
  Minneapolis, Minnesota. Association for Computational Linguistics.

\bibitem[{Feng et~al.(2021)Feng, Feng, Qin, Qin, and Liu}]{dialogBart}
Xiachong Feng, Xiaocheng Feng, Libo Qin, Bing Qin, and Ting Liu. 2021.
\newblock \href {https://doi.org/10.18653/v1/2021.acl-long.117} {Language model
  as an annotator: Exploring {D}ialo{GPT} for dialogue summarization}.
\newblock In \emph{Proceedings of the 59th Annual Meeting of the Association
  for Computational Linguistics and the 11th International Joint Conference on
  Natural Language Processing (Volume 1: Long Papers)}, pages 1479--1491,
  Online. Association for Computational Linguistics.

\bibitem[{Gliwa et~al.(2019)Gliwa, Mochol, Biesek, and Wawer}]{SAMSUM}
Bogdan Gliwa, Iwona Mochol, Maciej Biesek, and Aleksander Wawer. 2019.
\newblock \href {https://doi.org/10.18653/v1/D19-5409} {{SAMS}um corpus: A
  human-annotated dialogue dataset for abstractive summarization}.
\newblock In \emph{Proceedings of the 2nd Workshop on New Frontiers in
  Summarization}, pages 70--79, Hong Kong, China. Association for Computational
  Linguistics.

\bibitem[{Gu et~al.(2021)Gu, Yoo, and Ha}]{gu2021dialogbert}
Xiaodong Gu, Kang~Min Yoo, and Jung{-}Woo Ha. 2021.
\newblock \href {https://ojs.aaai.org/index.php/AAAI/article/view/17527}
  {Dialogbert: Discourse-aware response generation via learning to recover and
  rank utterances}.
\newblock In \emph{Thirty-Fifth {AAAI} Conference on Artificial Intelligence,
  {AAAI} 2021, Thirty-Third Conference on Innovative Applications of Artificial
  Intelligence, {IAAI} 2021, The Eleventh Symposium on Educational Advances in
  Artificial Intelligence, {EAAI} 2021, Virtual Event, February 2-9, 2021},
  pages 12911--12919. {AAAI} Press.

\bibitem[{Jacobs et~al.(1991)Jacobs, Jordan, Nowlan, and
  Hinton}]{jacobs1991adaptive}
Robert~A. Jacobs, Michael~I. Jordan, Steven~J. Nowlan, and Geoffrey~E. Hinton.
  1991.
\newblock \href {https://doi.org/10.1162/neco.1991.3.1.79} {Adaptive mixtures
  of local experts}.
\newblock \emph{Neural Comput.}, 3(1):79--87.

\bibitem[{Jia et~al.(2022)Jia, Liu, Tang, and Zhu}]{DiaSent}
Qi~Jia, Yizhu Liu, Haifeng Tang, and Kenny~Q. Zhu. 2022.
\newblock \href {https://doi.org/10.48550/arXiv.2204.13498} {Post-training
  dialogue summarization using pseudo-paraphrasing}.
\newblock \emph{CoRR}, abs/2204.13498.

\bibitem[{Lan et~al.(2020)Lan, Chen, Goodman, Gimpel, Sharma, and
  Soricut}]{lan2019albert}
Zhenzhong Lan, Mingda Chen, Sebastian Goodman, Kevin Gimpel, Piyush Sharma, and
  Radu Soricut. 2020.
\newblock \href {https://openreview.net/forum?id=H1eA7AEtvS} {{ALBERT:} {A}
  lite {BERT} for self-supervised learning of language representations}.
\newblock In \emph{8th International Conference on Learning Representations,
  {ICLR} 2020, Addis Ababa, Ethiopia, April 26-30, 2020}. OpenReview.net.

\bibitem[{Lewis et~al.(2020)Lewis, Liu, Goyal, Ghazvininejad, Mohamed, Levy,
  Stoyanov, and Zettlemoyer}]{lewis-etal-2020-bart}
Mike Lewis, Yinhan Liu, Naman Goyal, Marjan Ghazvininejad, Abdelrahman Mohamed,
  Omer Levy, Veselin Stoyanov, and Luke Zettlemoyer. 2020.
\newblock \href {https://doi.org/10.18653/v1/2020.acl-main.703} {{BART}:
  Denoising sequence-to-sequence pre-training for natural language generation,
  translation, and comprehension}.
\newblock In \emph{Proceedings of the 58th Annual Meeting of the Association
  for Computational Linguistics}, pages 7871--7880, Online. Association for
  Computational Linguistics.

\bibitem[{Li and Choi(2020)}]{li2020transformers}
Changmao Li and Jinho~D. Choi. 2020.
\newblock \href {https://doi.org/10.18653/v1/2020.acl-main.505} {Transformers
  to learn hierarchical contexts in multiparty dialogue for span-based question
  answering}.
\newblock In \emph{Proceedings of the 58th Annual Meeting of the Association
  for Computational Linguistics}, pages 5709--5714, Online. Association for
  Computational Linguistics.

\bibitem[{Li et~al.(2020{\natexlab{a}})Li, Liu, Kan, Zheng, Wang, Lei, Liu, and
  Qin}]{li2020molweni}
Jiaqi Li, Ming Liu, Min-Yen Kan, Zihao Zheng, Zekun Wang, Wenqiang Lei, Ting
  Liu, and Bing Qin. 2020{\natexlab{a}}.
\newblock \href {https://doi.org/10.18653/v1/2020.coling-main.238} {Molweni: A
  challenge multiparty dialogues-based machine reading comprehension dataset
  with discourse structure}.
\newblock In \emph{Proceedings of the 28th International Conference on
  Computational Linguistics}, pages 2642--2652, Barcelona, Spain (Online).
  International Committee on Computational Linguistics.

\bibitem[{Li et~al.(2021{\natexlab{a}})Li, Liu, Zheng, Zhang, Qin, Kan, and
  Liu}]{li2021dadgraph}
Jiaqi Li, Ming Liu, Zihao Zheng, Heng Zhang, Bing Qin, Min{-}Yen Kan, and Ting
  Liu. 2021{\natexlab{a}}.
\newblock \href {https://doi.org/10.1109/IJCNN52387.2021.9533364} {Dadgraph:
  {A} discourse-aware dialogue graph neural network for multiparty dialogue
  machine reading comprehension}.
\newblock In \emph{International Joint Conference on Neural Networks, {IJCNN}
  2021, Shenzhen, China, July 18-22, 2021}, pages 1--8. {IEEE}.

\bibitem[{Li et~al.(2020{\natexlab{b}})Li, Zhang, Zhao, Zhou, and
  Zhou}]{li2020task}
Junlong Li, Zhuosheng Zhang, Hai Zhao, Xi~Zhou, and Xiang Zhou.
  2020{\natexlab{b}}.
\newblock \href {http://arxiv.org/abs/2009.04984} {Task-specific objectives of
  pre-trained language models for dialogue adaptation}.
\newblock \emph{CoRR}, abs/2009.04984.

\bibitem[{Li et~al.(2022)Li, Wu, and Zhao}]{space}
Yiyang Li, Hongqiu Wu, and Hai Zhao. 2022.
\newblock \href {https://aclanthology.org/2022.coling-1.267}
  {Semantic-preserving adversarial code comprehension}.
\newblock In \emph{Proceedings of the 29th International Conference on
  Computational Linguistics}, pages 3017--3028, Gyeongju, Republic of Korea.
  International Committee on Computational Linguistics.

\bibitem[{Li and Zhao(2021)}]{sup}
Yiyang Li and Hai Zhao. 2021.
\newblock \href {https://doi.org/10.18653/v1/2021.findings-emnlp.176} {Self-
  and pseudo-self-supervised prediction of speaker and key-utterance for
  multi-party dialogue reading comprehension}.
\newblock In \emph{Findings of the Association for Computational Linguistics:
  EMNLP 2021}, pages 2053--2063, Punta Cana, Dominican Republic. Association
  for Computational Linguistics.

\bibitem[{Li et~al.(2021{\natexlab{b}})Li, Zhang, Fei, Feng, and
  Zhou}]{li-etal-2021-conversations}
Zekang Li, Jinchao Zhang, Zhengcong Fei, Yang Feng, and Jie Zhou.
  2021{\natexlab{b}}.
\newblock \href {https://doi.org/10.18653/v1/2021.acl-long.11} {Conversations
  are not flat: Modeling the dynamic information flow across dialogue
  utterances}.
\newblock In \emph{Proceedings of the 59th Annual Meeting of the Association
  for Computational Linguistics and the 11th International Joint Conference on
  Natural Language Processing (Volume 1: Long Papers)}, pages 128--138, Online.
  Association for Computational Linguistics.

\bibitem[{Liu et~al.(2021{\natexlab{a}})Liu, Zhang, Zhao, Zhou, and
  Zhou}]{liu2021filling}
Longxiang Liu, Zhuosheng Zhang, Hai Zhao, Xi~Zhou, and Xiang Zhou.
  2021{\natexlab{a}}.
\newblock \href {https://ojs.aaai.org/index.php/AAAI/article/view/17582}
  {Filling the gap of utterance-aware and speaker-aware representation for
  multi-turn dialogue}.
\newblock In \emph{Thirty-Fifth {AAAI} Conference on Artificial Intelligence,
  {AAAI} 2021, Thirty-Third Conference on Innovative Applications of Artificial
  Intelligence, {IAAI} 2021, The Eleventh Symposium on Educational Advances in
  Artificial Intelligence, {EAAI} 2021, Virtual Event, February 2-9, 2021},
  pages 13406--13414. {AAAI} Press.

\bibitem[{Liu et~al.(2019)Liu, Ott, Goyal, Du, Joshi, Chen, Levy, Lewis,
  Zettlemoyer, and Stoyanov}]{liu2019roberta}
Yinhan Liu, Myle Ott, Naman Goyal, Jingfei Du, Mandar Joshi, Danqi Chen, Omer
  Levy, Mike Lewis, Luke Zettlemoyer, and Veselin Stoyanov. 2019.
\newblock \href {https://arxiv.org/abs/1907.11692} {Roberta: A robustly
  optimized bert pretraining approach}.
\newblock \emph{arXiv preprint arXiv:1907.11692}.

\bibitem[{Liu et~al.(2021{\natexlab{b}})Liu, Feng, Wang, Song, Ren, and
  Zhang}]{liu2021graph}
Yongkang Liu, Shi Feng, Daling Wang, Kaisong Song, Feiliang Ren, and Yifei
  Zhang. 2021{\natexlab{b}}.
\newblock \href {https://ojs.aaai.org/index.php/AAAI/article/view/17585} {A
  graph reasoning network for multi-turn response selection via customized
  pre-training}.
\newblock In \emph{Thirty-Fifth {AAAI} Conference on Artificial Intelligence,
  {AAAI} 2021, Thirty-Third Conference on Innovative Applications of Artificial
  Intelligence, {IAAI} 2021, The Eleventh Symposium on Educational Advances in
  Artificial Intelligence, {EAAI} 2021, Virtual Event, February 2-9, 2021},
  pages 13433--13442. {AAAI} Press.

\bibitem[{Lowe et~al.(2015)Lowe, Pow, Serban, and Pineau}]{lowe2015ubuntu}
Ryan Lowe, Nissan Pow, Iulian Serban, and Joelle Pineau. 2015.
\newblock \href {https://doi.org/10.18653/v1/W15-4640} {The {U}buntu dialogue
  corpus: A large dataset for research in unstructured multi-turn dialogue
  systems}.
\newblock In \emph{Proceedings of the 16th Annual Meeting of the Special
  Interest Group on Discourse and Dialogue}, pages 285--294, Prague, Czech
  Republic. Association for Computational Linguistics.

\bibitem[{Qin et~al.(2021{\natexlab{a}})Qin, Gupta, Upadhyay, He, Choi, and
  Faruqui}]{qin2021timedial}
Lianhui Qin, Aditya Gupta, Shyam Upadhyay, Luheng He, Yejin Choi, and Manaal
  Faruqui. 2021{\natexlab{a}}.
\newblock \href {https://doi.org/10.18653/v1/2021.acl-long.549} {{TIMEDIAL}:
  Temporal commonsense reasoning in dialog}.
\newblock In \emph{Proceedings of the 59th Annual Meeting of the Association
  for Computational Linguistics and the 11th International Joint Conference on
  Natural Language Processing (Volume 1: Long Papers)}, pages 7066--7076,
  Online. Association for Computational Linguistics.

\bibitem[{Qin et~al.(2021{\natexlab{b}})Qin, Li, Che, Ni, and Liu}]{cogat}
Libo Qin, Zhouyang Li, Wanxiang Che, Minheng Ni, and Ting Liu.
  2021{\natexlab{b}}.
\newblock \href {https://ojs.aaai.org/index.php/AAAI/article/view/17616}
  {Co-gat: {A} co-interactive graph attention network for joint dialog act
  recognition and sentiment classification}.
\newblock In \emph{Thirty-Fifth {AAAI} Conference on Artificial Intelligence,
  {AAAI} 2021, Thirty-Third Conference on Innovative Applications of Artificial
  Intelligence, {IAAI} 2021, The Eleventh Symposium on Educational Advances in
  Artificial Intelligence, {EAAI} 2021, Virtual Event, February 2-9, 2021},
  pages 13709--13717. {AAAI} Press.

\bibitem[{Radford et~al.(2019)Radford, Wu, Child, Luan, Amodei, Sutskever
  et~al.}]{radford2019language}
Alec Radford, Jeffrey Wu, Rewon Child, David Luan, Dario Amodei, Ilya
  Sutskever, et~al. 2019.
\newblock Language models are unsupervised multitask learners.
\newblock \emph{OpenAI blog}, 1(8):9.

\bibitem[{Raffel et~al.(2020)Raffel, Shazeer, Roberts, Lee, Narang, Matena,
  Zhou, Li, and Liu}]{raffel2020exploring}
Colin Raffel, Noam Shazeer, Adam Roberts, Katherine Lee, Sharan Narang, Michael
  Matena, Yanqi Zhou, Wei Li, and Peter~J. Liu. 2020.
\newblock \href {http://jmlr.org/papers/v21/20-074.html} {Exploring the limits
  of transfer learning with a unified text-to-text transformer}.
\newblock \emph{J. Mach. Learn. Res.}, 21:140:1--140:67.

\bibitem[{Rajpurkar et~al.(2016)Rajpurkar, Zhang, Lopyrev, and
  Liang}]{rajpurkar2016squad}
Pranav Rajpurkar, Jian Zhang, Konstantin Lopyrev, and Percy Liang. 2016.
\newblock \href {https://doi.org/10.18653/v1/D16-1264} {{SQ}u{AD}: 100,000+
  questions for machine comprehension of text}.
\newblock In \emph{Proceedings of the 2016 Conference on Empirical Methods in
  Natural Language Processing}, pages 2383--2392, Austin, Texas. Association
  for Computational Linguistics.

\bibitem[{Reddy et~al.(2019)Reddy, Chen, and Manning}]{reddy2019coqa}
Siva Reddy, Danqi Chen, and Christopher~D. Manning. 2019.
\newblock \href {https://doi.org/10.1162/tacl_a_00266} {{C}o{QA}: A
  conversational question answering challenge}.
\newblock \emph{Transactions of the Association for Computational Linguistics},
  7:249--266.

\bibitem[{Sankar et~al.(2019)Sankar, Subramanian, Pal, Chandar, and
  Bengio}]{sankar-etal-2019-neural}
Chinnadhurai Sankar, Sandeep Subramanian, Chris Pal, Sarath Chandar, and Yoshua
  Bengio. 2019.
\newblock \href {https://doi.org/10.18653/v1/P19-1004} {Do neural dialog
  systems use the conversation history effectively? an empirical study}.
\newblock In \emph{Proceedings of the 57th Annual Meeting of the Association
  for Computational Linguistics}, pages 32--37, Florence, Italy. Association
  for Computational Linguistics.

\bibitem[{Smith et~al.(2020)Smith, Williamson, Shuster, Weston, and
  Boureau}]{smith-etal-2020-put}
Eric~Michael Smith, Mary Williamson, Kurt Shuster, Jason Weston, and Y-Lan
  Boureau. 2020.
\newblock \href {https://doi.org/10.18653/v1/2020.acl-main.183} {Can you put it
  all together: Evaluating conversational agents{'} ability to blend skills}.
\newblock In \emph{Proceedings of the 58th Annual Meeting of the Association
  for Computational Linguistics}, pages 2021--2030, Online. Association for
  Computational Linguistics.

\bibitem[{Sun et~al.(2019{\natexlab{a}})Sun, Yu, Chen, Yu, Choi, and
  Cardie}]{sun2019dream}
Kai Sun, Dian Yu, Jianshu Chen, Dong Yu, Yejin Choi, and Claire Cardie.
  2019{\natexlab{a}}.
\newblock \href {https://doi.org/10.1162/tacl_a_00264} {{DREAM}: A challenge
  data set and models for dialogue-based reading comprehension}.
\newblock \emph{Transactions of the Association for Computational Linguistics},
  7:217--231.

\bibitem[{Sun et~al.(2019{\natexlab{b}})Sun, Yu, Yu, and
  Cardie}]{sun-etal-2019-improving}
Kai Sun, Dian Yu, Dong Yu, and Claire Cardie. 2019{\natexlab{b}}.
\newblock \href {https://doi.org/10.18653/v1/N19-1270} {Improving machine
  reading comprehension with general reading strategies}.
\newblock In \emph{Proceedings of the 2019 Conference of the North {A}merican
  Chapter of the Association for Computational Linguistics: Human Language
  Technologies, Volume 1 (Long and Short Papers)}, pages 2633--2643,
  Minneapolis, Minnesota. Association for Computational Linguistics.

\bibitem[{Vaswani et~al.(2017)Vaswani, Shazeer, Parmar, Uszkoreit, Jones,
  Gomez, Kaiser, and Polosukhin}]{vaswani2017attention}
Ashish Vaswani, Noam Shazeer, Niki Parmar, Jakob Uszkoreit, Llion Jones,
  Aidan~N. Gomez, Lukasz Kaiser, and Illia Polosukhin. 2017.
\newblock \href
  {https://proceedings.neurips.cc/paper/2017/hash/3f5ee243547dee91fbd053c1c4a845aa-Abstract.html}
  {Attention is all you need}.
\newblock In \emph{Advances in Neural Information Processing Systems 30: Annual
  Conference on Neural Information Processing Systems 2017, December 4-9, 2017,
  Long Beach, CA, {USA}}, pages 5998--6008.

\bibitem[{Wu et~al.(2022)Wu, Ding, Zhao, Chen, Xie, Huang, and Zhang}]{whq}
Hongqiu Wu, Ruixue Ding, Hai Zhao, Boli Chen, Pengjun Xie, Fei Huang, and Min
  Zhang. 2022.
\newblock \href {https://doi.org/10.48550/ARXIV.2210.10293} {Forging multiple
  training objectives for pre-trained language models via meta-learning}.
\newblock \emph{CoRR}.

\bibitem[{Wu et~al.(2017)Wu, Wu, Xing, Zhou, and Li}]{wu2017sequential}
Yu~Wu, Wei Wu, Chen Xing, Ming Zhou, and Zhoujun Li. 2017.
\newblock \href {https://doi.org/10.18653/v1/P17-1046} {Sequential matching
  network: A new architecture for multi-turn response selection in
  retrieval-based chatbots}.
\newblock In \emph{Proceedings of the 55th Annual Meeting of the Association
  for Computational Linguistics (Volume 1: Long Papers)}, pages 496--505,
  Vancouver, Canada. Association for Computational Linguistics.

\bibitem[{Yang and Choi(2019)}]{yang2019friendsqa}
Zhengzhe Yang and Jinho~D. Choi. 2019.
\newblock \href {https://doi.org/10.18653/v1/W19-5923} {{F}riends{QA}:
  Open-domain question answering on {TV} show transcripts}.
\newblock In \emph{Proceedings of the 20th Annual SIGdial Meeting on Discourse
  and Dialogue}, pages 188--197, Stockholm, Sweden. Association for
  Computational Linguistics.

\bibitem[{Zhang et~al.(2020{\natexlab{a}})Zhang, Sun, Galley, Chen, Brockett,
  Gao, Gao, Liu, and Dolan}]{zhang-etal-2020-dialogpt}
Yizhe Zhang, Siqi Sun, Michel Galley, Yen-Chun Chen, Chris Brockett, Xiang Gao,
  Jianfeng Gao, Jingjing Liu, and Bill Dolan. 2020{\natexlab{a}}.
\newblock \href {https://doi.org/10.18653/v1/2020.acl-demos.30} {{DIALOGPT} :
  Large-scale generative pre-training for conversational response generation}.
\newblock In \emph{Proceedings of the 58th Annual Meeting of the Association
  for Computational Linguistics: System Demonstrations}, pages 270--278,
  Online. Association for Computational Linguistics.

\bibitem[{Zhang et~al.(2020{\natexlab{b}})Zhang, Sun, Galley, Chen, Brockett,
  Gao, Gao, Liu, and Dolan}]{DIALOGPT}
Yizhe Zhang, Siqi Sun, Michel Galley, Yen-Chun Chen, Chris Brockett, Xiang Gao,
  Jianfeng Gao, Jingjing Liu, and Bill Dolan. 2020{\natexlab{b}}.
\newblock \href {https://doi.org/10.18653/v1/2020.acl-demos.30} {{DIALOGPT} :
  Large-scale generative pre-training for conversational response generation}.
\newblock In \emph{Proceedings of the 58th Annual Meeting of the Association
  for Computational Linguistics: System Demonstrations}, pages 270--278,
  Online. Association for Computational Linguistics.

\bibitem[{Zhang et~al.(2018)Zhang, Li, Zhu, Zhao, and Liu}]{zhang2018modeling}
Zhuosheng Zhang, Jiangtong Li, Pengfei Zhu, Hai Zhao, and Gongshen Liu. 2018.
\newblock \href {https://aclanthology.org/C18-1317} {Modeling multi-turn
  conversation with deep utterance aggregation}.
\newblock In \emph{Proceedings of the 27th International Conference on
  Computational Linguistics}, pages 3740--3752, Santa Fe, New Mexico, USA.
  Association for Computational Linguistics.

\bibitem[{Zhang et~al.(2021)Zhang, Li, and Zhao}]{zhang2021multi}
Zhuosheng Zhang, Junlong Li, and Hai Zhao. 2021.
\newblock \href {https://doi.org/10.1109/TASLP.2021.3058616} {Multi-turn
  dialogue reading comprehension with pivot turns and knowledge}.
\newblock \emph{{IEEE} {ACM} Trans. Audio Speech Lang. Process.},
  29:1161--1173.

\bibitem[{Zhou et~al.(2018)Zhou, Li, Dong, Liu, Chen, Zhao, Yu, and Wu}]{dam}
Xiangyang Zhou, Lu~Li, Daxiang Dong, Yi~Liu, Ying Chen, Wayne~Xin Zhao, Dianhai
  Yu, and Hua Wu. 2018.
\newblock \href {https://doi.org/10.18653/v1/P18-1103} {Multi-turn response
  selection for chatbots with deep attention matching network}.
\newblock In \emph{Proceedings of the 56th Annual Meeting of the Association
  for Computational Linguistics (Volume 1: Long Papers)}, pages 1118--1127,
  Melbourne, Australia. Association for Computational Linguistics.

\end{thebibliography}
\bibliographystyle{acl_natbib}

\clearpage
\appendix
\section{Hyper-parameter Settings}
\label{app:hyper}
In this section, we present the detailed hyper-parameter settings of each dataset.
\subsection{Hyper-parameters for MuTual}
For both MuTual and MuTual$\operatorname{^{plus}}$, we set the maximum input sequence length to 320, where the maximum response length is set to 52 which means the maximum dialogue history length is 268. When truncating the input sequence, we only truncate the dialogue history and leave the response candidates intact. To guarantee the fluency of dialogue history, we truncate them from the front, and at the unit of utterances instead of tokens. The learning rate, training epochs, and batch size are set to 6e-6, 3, and 2, respectively. We use AdamW as our training optimizer and a linear scheduler to schedule the learning rate. The learning rate is first linearly warmed up from 0 to 6e-6 at the first 1\% steps then decreased linearly to 0 until the end of training.

\subsection{Hyper-parameters for Molweni}
For the Molweni dataset, the maximum input sequence length is set to 384, where the maximum question length is 32. Similar to the MuTual dataset, we only truncate the dialogue history and leave the question sentence intact. The learning rate, training epochs, and batch size are set to 7e-5, 5, and 16, respectively. As for the optimizer and scheduler, they are the same as the ones on MuTual dataset.

\subsection{Hyper-parameters for DIALOGSUM}
For the DIALOGSUM dataset, the maximum input sequence length and maximum summary length are set to 512 and 100, respectively. The learning rate, training epochs, and batch size are set to 2e-5, 15, and 12, respectively. During inference, we use beam search to generate summaries, and set the beam size to 4.

\section{Results on SAMSum Dataset}
\label{app:samsum}
For the dialogue summarization task, we also conduct experiments on the SAMSum \cite{SAMSUM} dataset. SAMSum is a dialogue summarization dataset that contains 16,369 dialogues in the form of online chatting messages. Compared with DIALOGSUM, which is taken from real-life person-to-person conversations, this dataset contains dialogues that are more informal and colloquial. However, the summaries in this dataset are less abstractive than DIALOGSUM \cite{chen2021dialsumm}.

Results on SAMSum are tabulated in Table \ref{tab_samsum}, where we can see that BiDeN consistently outperforms the strong baseline BART by large margins. We also compare BiDeN with different models that are also built on BART. Multi-View BART \cite{multi-view} incorporates different information like topic and stage of dialogues to generate summaries using a multi-view decoder. ConDigSum is the current SOTA model on the SAMSum dataset, which designs two contrastive auxiliary tasks: Coherence Detection and Sub-summary Generation to implicitly model the topic information of dialogues. This model is trained with an alternating updating strategy, which is approximately three times slower than our BiDeN during training since it requires three backward calculations in a single batch. DialoBART and DialSent-PGG are introduced in Section \ref{sec:result_dialogsum}. Table \ref{tab_samsum} shows that BiDeN achieves comparable results to ConDigSum and outperforms all other models. It is worth noting that all of the previous models require additional dialogue annotators or training stages, while our BiDeN is annotator-free, plug-and-play, and easy to use.

Note that the original results of Multi-View and ConDigSum are obtained by the files2rouge package based on the official ROUGE-1.5.5.pl Perl script, while DialoBART and DialSent-PGG adopt py-rouge. To make fair comparisons, we download the output predictions of Multi-View and ConDigSum, then run the py-rouge script to get the corresponding results, which are the ones presented in Table \ref{tab_samsum}.

For the SAMSum dataset, we set the maximum dialogue history length to 800, and the maximum summary length to 100. The learning rate, training epochs, and batch size are set to 2e-5, 5, and 4, respectively. We also adopt beam search during inference, where the beam size is also set to 4.

\begin{table}[tbp]
    \centering
    \small
    \begin{tabular}{l r r r}
        \specialrule{0.09em}{0.0pt}{2.0pt}
        Model & \textbf{Rouge-1} & \textbf{Rouge-2} & \textbf{Rouge-L}\\
        \specialrule{0.03em}{1.3pt}{0.5pt}
        \specialrule{0.03em}{0.5pt}{1.3pt}
        Multi-View BART & 0.534 & 0.280 & 0.499\\
        DialSent-PGG & 0.535 & 0.289 & 0.502\\
        DialoBART & 0.537 & 0.288 & 0.508 \\
        ConDigSum & \textbf{0.542} & 0.289 & \textbf{0.509} \\
        BART & 0.526 & 0.271 & 0.492 \\
        \quad +BIDM & $^*$0.531 & $^*$0.278 & $^{*}$0.498\\
        \quad \quad +BiDeN & $^{**}$0.540 & $^{**}$\textbf{0.291} & $^{**}$0.506\\
        \specialrule{0.09em}{1.2pt}{0.0pt}
    \end{tabular}
    \caption{Results on SAMSum, where $^*$ and $^{**}$ represent the same as in Table \ref{tab_mutual_dev}.}
    \label{tab_samsum}
\end{table}

\section{More Visualizations}
\label{app:more_vis}
We present more examples of the three kinds of attentions: current-to-past attention, current-to-future attention, and current-to-current attention, for readers to further explore how BiDeN works.

\begin{figure}[tbp]
	\centering
	\subfigure[Current-to-past Attention]{
		\begin{minipage}{0.46\textwidth} 
            \includegraphics[width=0.96\textwidth]{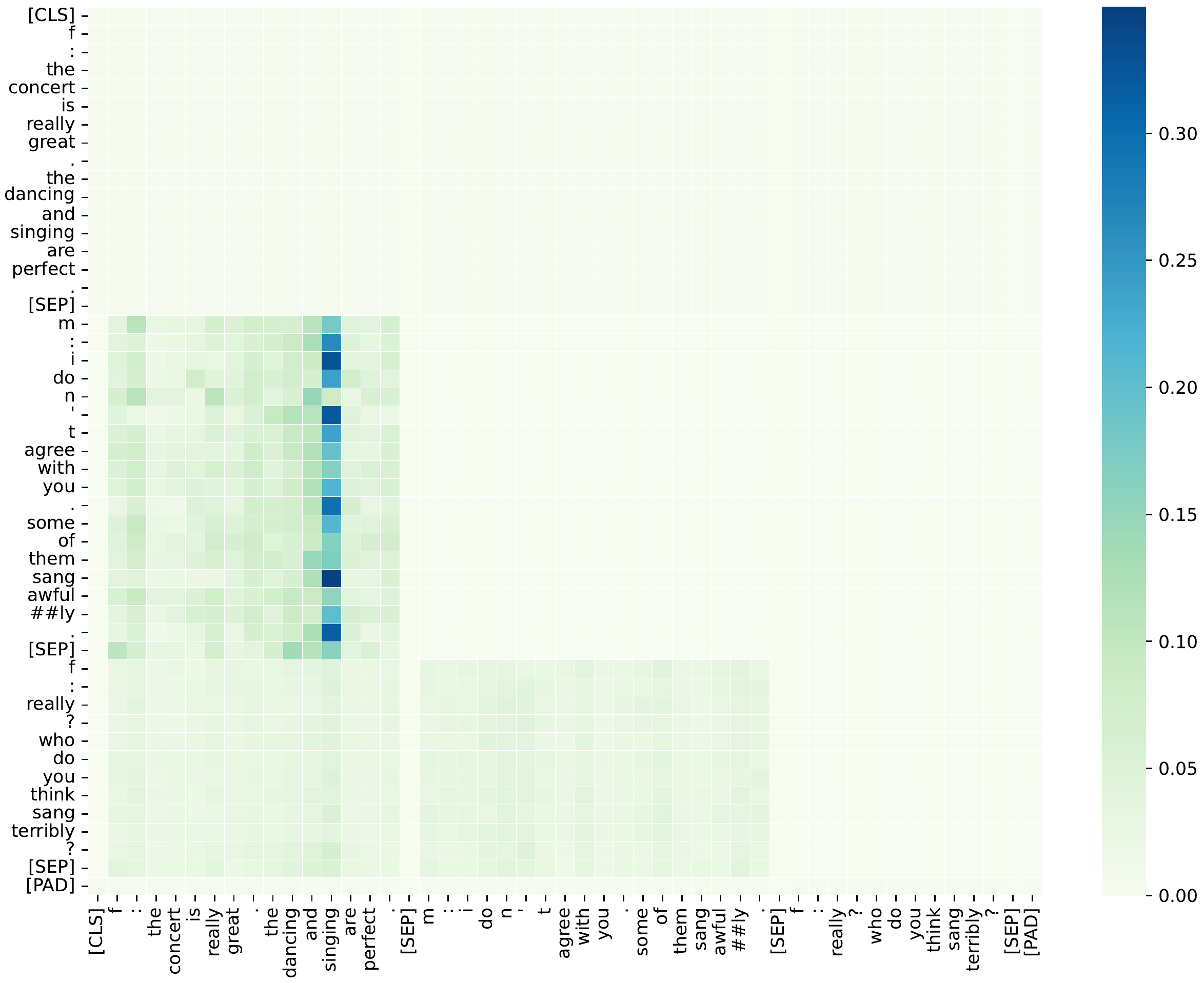} \\
		\end{minipage}
	}

	\subfigure[Current-to-future Attention]{
		\begin{minipage}{0.46\textwidth} 
            \includegraphics[width=0.96\textwidth]{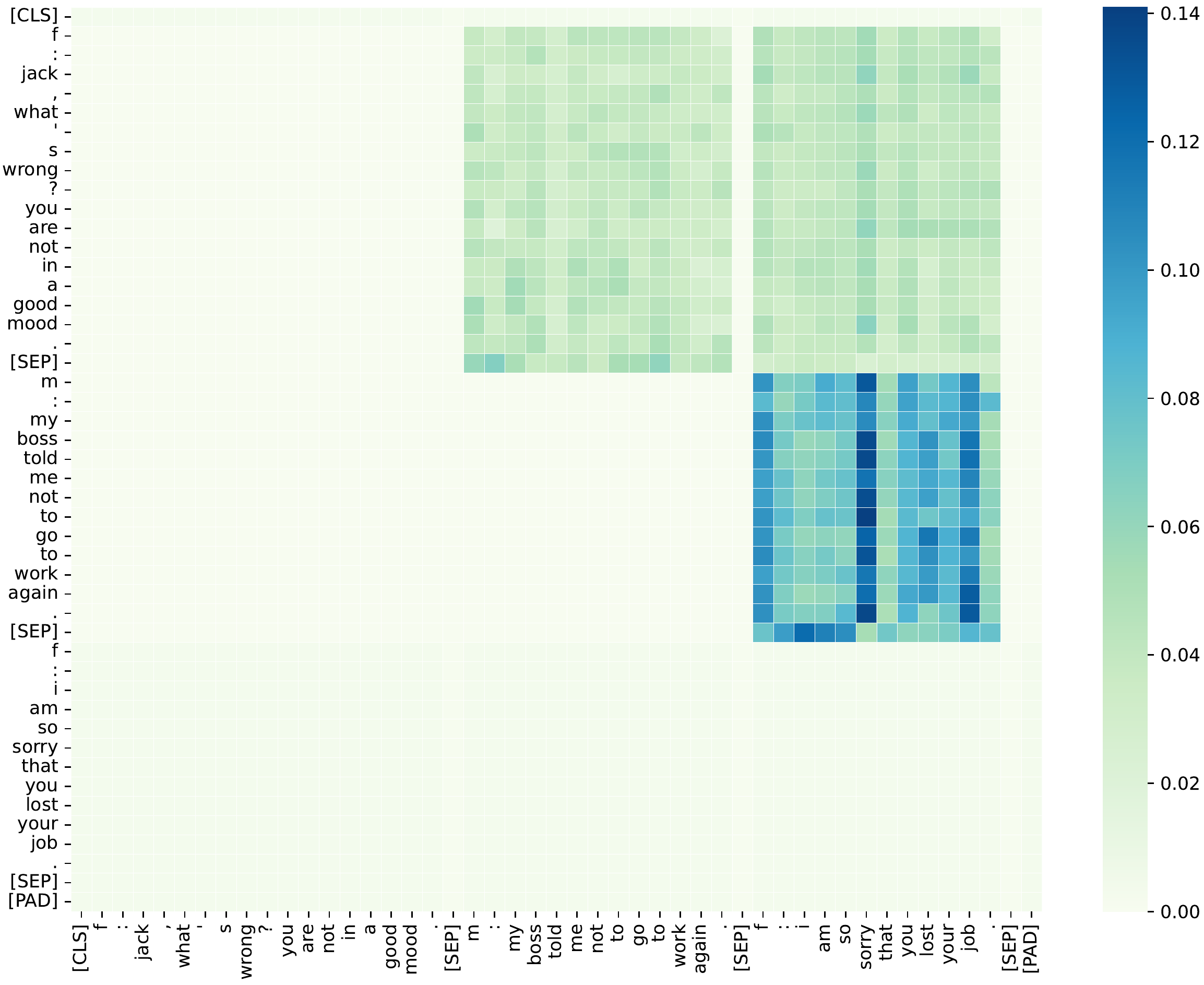} \\
		\end{minipage}
	}
	
	\subfigure[Current-to-current Attention]{
		\begin{minipage}{0.46\textwidth} 
            \includegraphics[width=0.96\textwidth]{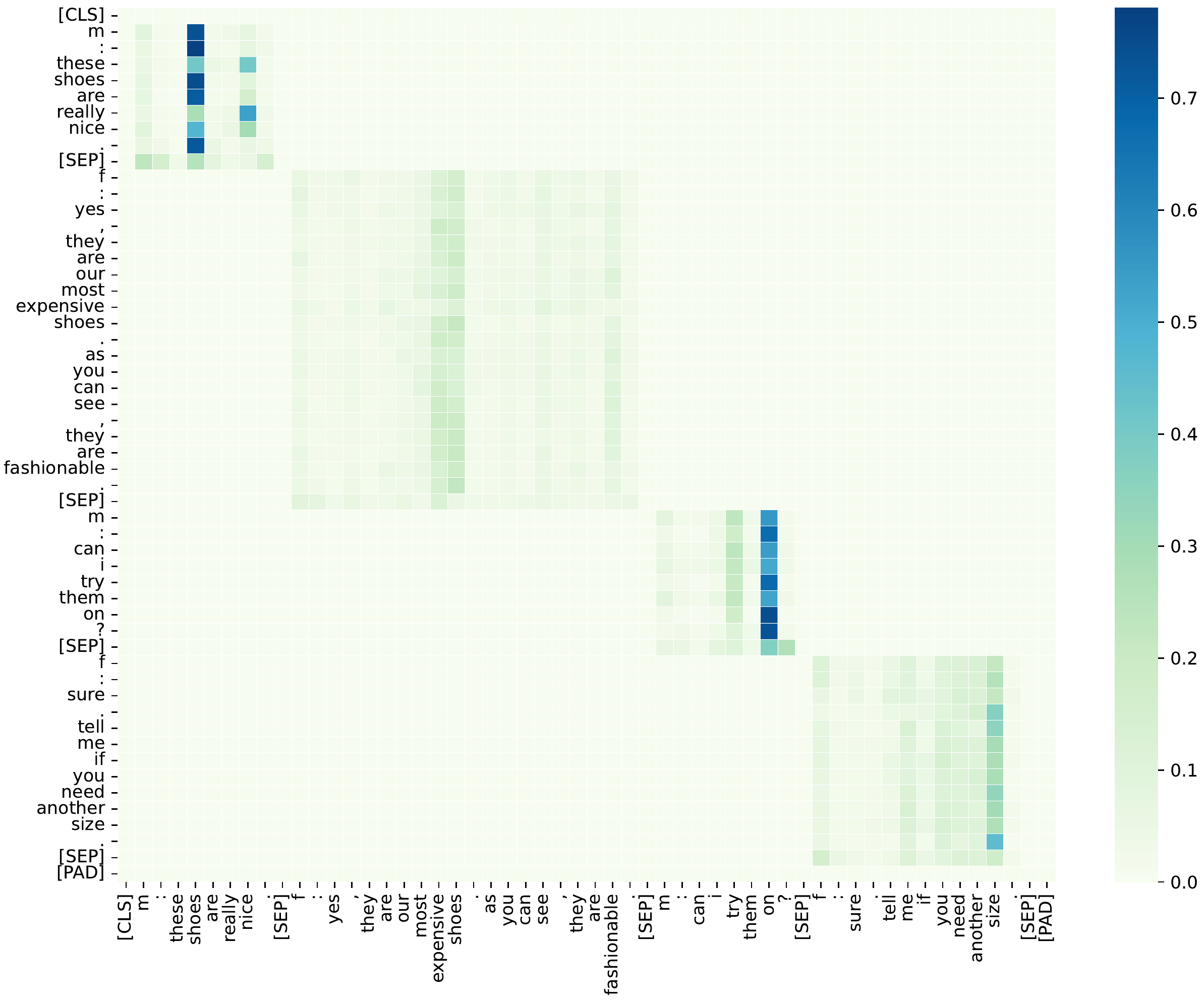} \\
		\end{minipage}
	}

	\caption{More visualization results.}
	\label{fig_more_attn}
\end{figure}

Figure \ref{fig_more_attn} (a) illustrates a conversation about a concert, where the female thinks the dancing and singing are perfect but the male disagrees. We can see from the attention weights that when modeling the second utterance, BiDeN focuses mostly on \emph{dancing and singing}, especially on \emph{singing}, which is consistent with its semantic meaning that some singers sang awfully. In other words, BiDeN is capable of extracting the key information of previous utterances when modeling the current utterance.

Figure \ref{fig_more_attn} (b) is another example of Current-to-future attention, where the male is unhappy because he lost his job and the female feels sorry about that. It can be observed that when modeling the second utterance, BiDeN attends more on \emph{sorry} and \emph{you lost your job}. This observation demonstrates that BiDeN is able to locate the key information in the future utterances to model
what kind of current utterance will lead to the development of the future dialogue.

Figure \ref{fig_more_attn} (c) shows an example of current-to-current attention, which is the self-attention within each utterance. Let's focus on each utterance. The first utterance mainly attends to \emph{shoes} and \emph{nice}, which are two keywords that best reflect the semantic meaning of this utterance. Similar observations can be seen in the rest three utterances, where the most prominent words are \emph{expensive shoes} and \emph{fashionable}, \emph{try on}, and \emph{you need another size}, respectively. This observation indicates that BiDeN can model the most salient and concise semantic meaning in each utterance.

\end{document}